\documentclass[final,3p,numbers,nopreprintline]{elsarticle}

\usepackage{lmodern}
\usepackage{amsmath,amssymb}
\usepackage{microtype}
\usepackage{booktabs}
\usepackage{array}
\usepackage{flafter}
\usepackage{wrapfig}
\usepackage[hidelinks]{hyperref}

\journal{Journal of Computational Physics}
\biboptions{sort&compress}
\graphicspath{{figures/}}
\AddToHook{env/table/begin}{%
  \setlength{\abovecaptionskip}{5pt}%
  \setlength{\belowcaptionskip}{5pt}}

\newcommand{\method}{SIDNO}
\newcommand{\methodplus}{SIDNO+}
\newcommand{\tabledash}{\textendash\llap{\rule[0.22em]{0.5em}{0.09em}}}
\newcommand{\vu}{\mathbf{u}}
\newcommand{\vM}{\mathbf{M}}
\newcommand{\vB}{\mathbf{B}}
\newcommand{\vn}{\mathbf{n}}
\newcommand{\vI}{\mathbf{I}}
\IfFileExists{nicefrac.sty}{\usepackage{nicefrac}}{\usepackage{xfrac}\providecommand{\nicefrac}{\sfrac}}
\newcommand{\halfdt}{\nicefrac{\Delta t}{2}}

\hypersetup{pdftitle={Separating Ideal and Dissipative Dynamics in a Neural Operator for Non-Ideal Magnetohydrodynamics},pdfsubject={Manuscript for the Journal of Computational Physics}}

\begin{document}
\begin{frontmatter}
\title{Separating Ideal and Dissipative Dynamics in a Neural Operator for Non-Ideal Magnetohydrodynamics}
\author[brain,miriai]{Dmitry Bylinkin}
\author[harbin-lab,harbin-center]{Xiaowei Jin}
\author[brain,miriai,innopolis]{Aleksandr Beznosikov}
\author[harbin-lab,harbin-center]{Hui Li}

\affiliation[brain]{op={},organization={Basic Research of Artificial Intelligence Laboratory (BRAIn Lab)}}
\affiliation[miriai]{op={},organization={Moscow Independent Research Institute of Artificial Intelligence}}
\affiliation[harbin-lab]{op={},organization={Key Lab of Structures Dynamic Behavior and Control of the Ministry of Education, Harbin Institute of Technology, Harbin 150090, China}}
\affiliation[harbin-center]{op={},organization={International Research Center for Intelligent Fluid Mechanics, Harbin Institute of Technology, Harbin 150090, China}}
\affiliation[innopolis]{op={},organization={Innopolis University}}

\begin{abstract}
In compressible, isothermal magnetohydrodynamics, ideal transport and dissipation change the same fields, making their contributions difficult for a learnable mapping to distinguish.
We introduce the Strang Ideal-Dissipative Neural Operator (SIDNO), which places a learned ideal update between numerical half-steps for viscosity and magnetic diffusion.
The ideal branch predicts potentials that a conservative decoder uses to update density, momentum and magnetic flux while preserving total mass and discrete magnetic divergence.
In an unobstructed periodic domain and in flows around permeable bodies, the ideal update contributes little to the reference free-energy loss.
On body-fitted meshes, SIDNO achieves lower rollout error than the competitors at comparable computational cost and retains its advantage for unseen body shapes, changes in topology and Reynolds numbers above the training range.
Distilling the dissipative half-step into a local learned update retains most of the reference dissipation in both settings and shortens rollouts on finely resolved body-fitted meshes.
\end{abstract}

\end{frontmatter}

\section{Introduction}
\label{sec:intro}

Applications of mathematical physics may require solving the same partial differential equations (PDEs) for many candidate geometries or operating conditions \citep{subbarao2009combined}.
Neural operators learn the solution map of a PDE from simulation data and replace the solver at a fixed cost per step \citep{kovachki2023neural}.
This replacement is most useful where the numerical solver is expensive.
Magnetized flows are one such case \citep{pitta2018predicting}.
Magnetohydrodynamics (MHD) couples the flow and the magnetic field through nonlinear terms.
Viscosity and resistivity add stiff diffusive terms and set the width of thin shear layers and current sheets, which a solver has to resolve with fine meshes and small time steps.
Such flows arise around bodies in space plasmas \citep{harnett2000two,jia2009properties,corso2026simulations} and in engineering settings, where magnetized flows pass obstacles \citep{burdiak2017structure} or applied fields control hypersonic flow \citep{poggie2002magnetic,muir2022numerical}.

Operator learning for MHD started with incompressible and reduced models.
\citet{rosofsky2023magnetohydrodynamics} trained physics-informed neural operators on incompressible resistive MHD, \citet{gopakumar2024plasma} used the Fourier neural operator (FNO) \citep{li2020fourier} as a surrogate of reduced MHD in tokamak plasmas, and \citet{kim2025neural} showed that neural operators learn local flux updates of ideal MHD.
The Well \citep{ohana2024well} added compressible isothermal MHD turbulence to a public benchmark, and later models were trained on these data \citep{shikhman2026cellular}, but the equations there are ideal.
Non-ideal MHD, however, is of greater practical interest and is at the same time harder to learn.
Indeed, in this case the operator has to capture two kinds of dynamics at once: the ideal part carries sound and Alfv\'en waves, while the dissipative part smooths the fields at rates set by the viscosity and the resistivity.
To learn the ideal part, the operator has to track the fluxes of mass, momentum and magnetic field, but viscosity and magnetic diffusion change the same fluxes, with a strength that varies across the data.
A single network represents both parts with the same weights, and a loss on the next state does not tell it which part of a predicted increment is transport and which is dissipation.
Without an inductive bias, the network has to infer how those two contributions combine from observations of the full dynamics.
As a result, it is harder for the operator to generalize along coupled axes.

A natural inductive bias is to build the split between ideal and dissipative dynamics into the architecture.
The operator then has two branches: one learns the ideal dynamics, and the other reproduces the viscous and resistive dissipation \citep{sulskis2026generic,deveney2026elucidating}.
Each branch generalizes along its own axes, and the training no longer has to sort transport from dissipation inside a single network.
Existing neural operator architectures separate the dynamics into conservative and dissipative parts using the metriplectic formalism \citep{morrison1986paradigm}.
A skew-symmetric operator acting on the energy gradient generates a flow that conserves energy, and a symmetric positive semidefinite operator acting on the entropy gradient generates dissipation by producing entropy.
\citet{hernandez2021structure} learn two operators independently of one another, together with the energy and entropy, and penalize violations of the degeneracy conditions during training.
\citet{lee2021machine} instead parameterize the two operators to satisfy the degeneracy conditions identically, making energy conservation and nondecreasing entropy properties of the learned dynamics for any weights.
\citet{zhang2022gfinns} and \citet{gruber2025efficiently} reduce the cost of such parameterizations and scale them to larger states.
\citet{park2024tlasdi} learn a reduced representation and impose the same structure on the evolution in the latent space.
\citet{sulskis2026generic} bring this structure into an FNO, and \citet{baheri2025metriplectic} use it in a flow-matching generative model of dissipative dynamics.
In all of these models both branches are learned jointly from trajectories.
The trajectory loss constrains the combined evolution, and thus different branches can give the same fitted dynamics \citep{sulskis2026generic}.
As a result, the conservative branch need not conserve the physical energy, and the dissipative branch need not reproduce the physical viscous and resistive losses.
This leaves open \textit{which architecture and training pipeline can assign the two kinds of dynamics to their respective branches}.

The metriplectic neural operators mentioned above were developed for PDEs on regular grids.
Applied problems, in contrast, involve obstacles and thin layers and therefore require irregular meshes that resolve their surfaces.
MeshGraphNets \citep{pfaff2020learning} pass messages along mesh edges, and the receptive field grows with the number of message-passing steps.
GINO \citep{li2023geometry} uses graph neural operators to lift the input to a regular latent grid and to evaluate the output at query points, with an FNO acting on the latent grid.
Transolver \citep{wu2024transolver} groups mesh points into learned slices according to their physical states and applies attention across slices.
UPT \citep{alkin2024universal} compresses the input into a fixed set of latent tokens for propagation, and OFormer \citep{li2022transformer} uses cross-attention to decode an encoded point set at arbitrary query positions.
DoMINO \citep{ranade2025domino} represents the geometry around each query point with local neighborhoods at several scales.
These general-purpose architectures are well studied for engineering problems, with few adaptations to MHD. Thus, the second open question is \textit{how to implement the ideal and dissipative branches on complex irregular domains}.

To our knowledge, no existing neural operator separates ideal and dissipative dynamics and generalizes well for compressible non-ideal MHD.
This paper makes the following contributions.
\begin{itemize}
\item We propose the Strang Ideal-Dissipative Neural Operator (\method{}).
Each step is a symmetric sequence of a dissipative half-step, a conservative ideal step and a second dissipative half-step.
This Strang order \citep{strang1968construction} keeps the error of the splitting at second order in the time step and lets the ideal step act on a state that has already passed half of the dissipation.
A conservative decoder preserves mass and the divergence of the magnetic field. The two dissipative half-steps account for a median of $99.9\%$ of the reference energy loss per step in unobstructed periodic flows and $101.1\%$ on the finite-element test set. On both test sets the ideal step changes the energy by a median of $-1.9\%$ of that loss.
\item We propose a two-stage training pipeline for \method{}.
The operator is trained with the exact dissipative step as a teacher, and this step is then distilled into a cheap one-pass student.
The learned dissipative step accounts for a median of $99.1\%$ of the reference energy loss per step on the regular grid and $92.9\%$ on body-fitted meshes.
\item We generate two datasets of compressible isothermal MHD: 3047 trajectories on a regular grid and 2431 trajectories on finite-element meshes around permeable bodies.
Both datasets vary the viscosity, resistivity and initial sonic and Alfv\'en Mach numbers. In flows with a body we also vary its shape and permeability. The test sets extend the Reynolds numbers beyond the training range and, for flows around bodies, add unseen shapes and topologies.
\item In the unobstructed periodic box, \method{} reaches a normalized root mean squared error (NRMSE) of $0.1056$ after 50 rollout steps on the in-distribution test set, compared with $0.2441$ for the closest measured baseline at a comparable cost per step. For flows around permeable bodies, the corresponding errors are $0.0989$ and $0.2187$.
\method{} retains its advantage over the strongest in-distribution baseline for unseen body shapes, flows around two bodies, and Reynolds numbers above the training range.
\end{itemize}

\section{Methodology}
\label{sec:method}

\subsection{Problem Formulation}
\label{sec:problem}

Let $\Omega$ be a periodic box. In the unobstructed case the fluid fills $\Omega$ and $\Omega_b=\varnothing$. When one or more stationary rigid bodies with interior $\Omega_b$ are present, the fluid occupies $\Omega_f=\Omega\setminus\overline{\Omega_b}$, and the unit normal $\vn$ on the common interface $\Gamma=\partial\Omega_b$ points from the body into the fluid.
In $\Omega_f$, the density $\rho$, momentum $\vM=\rho\vu$ and magnetic field $\vB_f$ satisfy the compressible isothermal MHD equations:
\begin{subequations}
\label{eq:mhd}
\begin{equation}
\begin{aligned}
&\partial_t\rho + \nabla\cdot\vM = 0,\\
&\partial_t\vM + \nabla\cdot\Big(\frac{\vM\otimes\vM}{\rho} + c_s^2\rho\,\vI\Big)
  + \vB_f\times(\nabla\times\vB_f) = \nabla\cdot\big(2\nu\rho\,S(\vu)\big),\\
&\partial_t\vB_f = \nabla\times(\vu\times\vB_f)+\eta\,\nabla^2\vB_f,
  \qquad \nabla\cdot\vB_f=0.
\end{aligned}
\label{eq:mhd-fluid}
\end{equation}
Here $c_s$ is the sound speed, $\nu$ is the kinematic viscosity, $\eta$ is the magnetic diffusivity, and $S(\vu)=\nicefrac{1}{2}(\nabla\vu+\nabla\vu^\top)-\nicefrac{1}{3}(\nabla\cdot\vu)\,\vI$ is the traceless strain rate.
In $\Omega_b$, the magnetic field $\vB_b$ evolves by diffusion with diffusivity $\eta_b$,
\begin{equation}
\partial_t\vB_b=\eta_b\,\nabla^2\vB_b,\qquad \nabla\cdot\vB_b=0.
\label{eq:mhd-body}
\end{equation}
On $\Gamma$, the fluid satisfies no slip, and the two magnetic fields are coupled by continuity of normal flux and of the tangential magnetic field strength:
\begin{equation}
\left.\vu\right|_\Gamma=0,\qquad
\vn\cdot(\vB_f-\vB_b)=0,\qquad
\vn\times\big(\vB_f-\nicefrac{\vB_b}{\mu_b}\big)=0,
\label{eq:mhd-interface}
\end{equation}
\end{subequations}
where $\mu_b$ is the permeability in the body, and we assume unit permeability ($\mu=1$) in the fluid. We write $\vB$ for the field obtained from $\vB_f$ and $\vB_b$ and $\xi=(\rho,\vM,\vB)$ for the state. The initial density, momentum and magnetic field are prescribed, with periodic conditions at the outer boundary of the box.
The flow is characterized by the sonic and Alfv\'en Mach numbers $M_s=\nicefrac{u_{\mathrm{rms}}}{c_s}$ and $M_A=\nicefrac{u_{\mathrm{rms}}}{v_A}$, where $v_A$ is the root-mean-square Alfv\'en speed $\nicefrac{|\vB|}{\sqrt{\rho}}$, and by the Reynolds and magnetic Reynolds numbers $\mathrm{Re}=\nicefrac{u_{\mathrm{rms}}L_{\mathrm{int}}}{\nu}$ and $\mathrm{Rm}=\nicefrac{u_{\mathrm{rms}}L_{\mathrm{int}}}{\eta}$ with the integral length scale $L_{\mathrm{int}}$ of the flow.

\subsection{Core Architecture}
\label{sec:core}

\begin{figure}[!t]
\centering
\includegraphics[width=\linewidth]{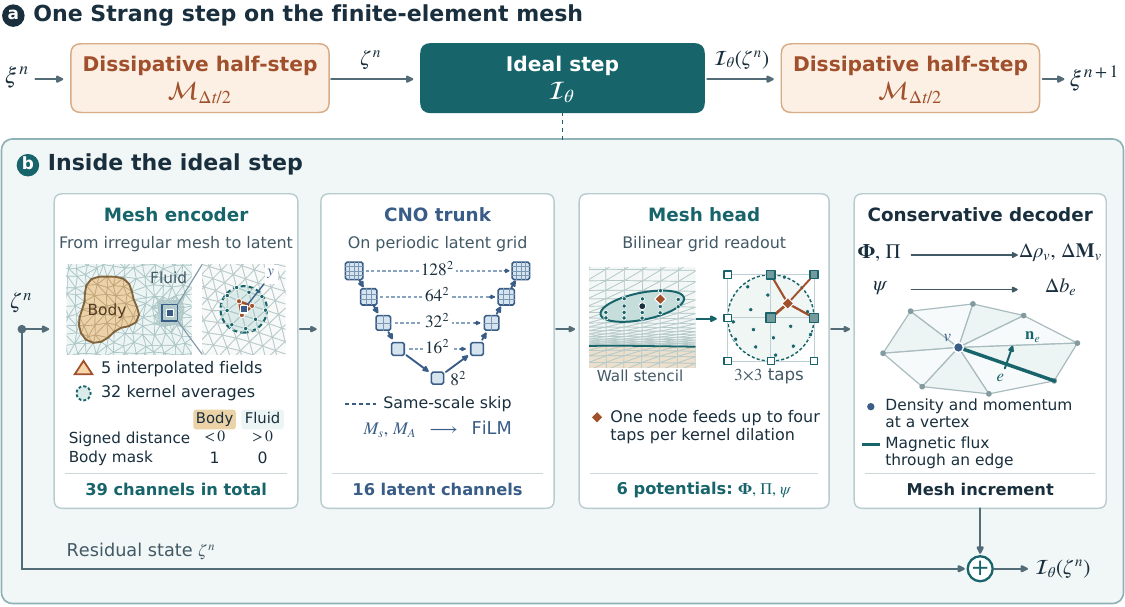}
\caption{One step of \method{}. (a)~Strang splitting places the ideal step between two dissipative half-steps on the mesh. (b)~The encoder maps the intermediate state $\zeta^n$ to a regular latent grid, where the CNO trunk advances it. The mesh head predicts potentials at fluid nodes, and the conservative decoder converts them into the increment added to $\zeta^n$.}
\label{fig:architecture}
\end{figure}

The two discretizations use the same splitting. On the regular grid, the ideal branch acts directly on grid fields and the conservative decoder takes periodic derivatives of its predicted potentials. On body-fitted meshes, \method{} represents density and momentum at the fluid nodes, and the magnetic field is represented by its fluxes through the edges of the mesh, with zero discrete divergence on every triangle. Figure~\ref{fig:architecture} shows this mesh implementation.
We split the terms of \eqref{eq:mhd} into an ideal part, which holds the transport terms, and a dissipative part, which holds the viscous stress and the magnetic diffusion.
To assign the two parts to separate layers of the model, we use the Strang splitting \citep{strang1968construction}:
\begin{equation}
\xi^{n+1}=\mathcal M_{\halfdt}\circ\mathcal I_\theta\circ\mathcal M_{\halfdt}\,(\xi^n), \label{eq:strang}
\end{equation}
where $\mathcal M_{\halfdt}$ advances the dissipative part over half a step and $\mathcal I_\theta$ approximates the evolution under the ideal part over $\Delta t$.
We write $\zeta^n=\mathcal M_{\halfdt}(\xi^n)$ for the input to the ideal step.
We place the dissipative part outside because the ideal part is harder to learn.
The learned ideal step receives a state that has already passed half of the dissipation, and the second dissipative half-step damps the small-scale errors of its increment before the next step.

The density stays fixed during the dissipative half-step, and the dissipative terms of \eqref{eq:mhd} form a linear system $\partial_t(\vM,\vB)=\mathcal D_\rho(\vM,\vB)$ with known coefficients.
For momentum, $\mathcal D_\rho$ gives the divergence of the viscous stress $2\nu\rho\,S(\vM/\rho)$ in the fluid. For the magnetic field, it gives the resistive diffusion rates $\eta\nabla^2\vB_f$ in the fluid and $\eta_b\nabla^2\vB_b$ in the body, subject to the interface conditions in \eqref{eq:mhd-interface}.
The half-step is therefore the exponential $\mathcal M_{\halfdt}=\exp\big((\halfdt)\mathcal D_\rho\big)$.
Its sparse matrices are known before the step, and evaluating their action requires a series of sparse products without nonlinear iterations.
The ideal equations, on the other hand, are nonlinear and hyperbolic, hence a solver has to use implicit time stepping to exceed the explicit stability limit and then solve a nonlinear system at every step.
The non-learned half-step gives the model a way to separate the two parts, since the layers of $\mathcal I_\theta$ do not have to learn the effects of viscosity and resistivity on the flow.
The half-step, in turn, never has to know the nonlinear dynamics of the system, and layers with a small number of learned parameters can therefore replace it.
Section~\ref{sec:learnable} makes the dissipative half-step learnable via distillation.
\ref{app:dissipative} gives the finite-element discretization and the evaluation of the exponential.

To move the input of the ideal step to a periodic latent grid, we first form nodal fields.
Each node also carries the distance to $\Gamma$ in units of the local mesh size, the wall normal and the orientation of the local mesh anisotropy. We denote these features by $f_v$.
The encoder output at a grid point $y$ stacks three groups of channels.
Five channels sample the nodal fields at $y$ by linear interpolation on the triangle that contains $y$, and grid points inside the body take the values at the nearest point of $\Gamma$.
Two channels are the signed distance to $\Gamma$ and the body mask.
The remaining channels come from a learned kernel integral over the fluid nodes $x_v$ near $y$,
\begin{equation}
E_\theta(y)=\sum_{v:\,|x_v-y|<r}\omega_{yv}\,\kappa_\theta\Big(\frac{x_v-y}{r},\,f_v\Big)\odot W\big[\zeta_v,\,f_v\big],\qquad \omega_{yv}=\frac{A_v}{\sum_{k:\,|x_k-y|<r}A_k}, \label{eq:encoder}
\end{equation}
where $r$ is the radius of two grid cells, $W$ is a linear lift, $\zeta_v$ is the vector of nodal fields at $x_v$ and $\kappa_\theta$ is a small network.
The dual area $A_v$ is one third of the sum of the areas of the fluid triangles incident to $v$. The normalized weights $\omega_{yv}$ make the sum approximate an area mean over the disk even where the mesh is refined near $\Gamma$.
\ref{app:encoder} gives the construction and normalization of the encoder inputs.

A convolutional neural operator (CNO) \citep{raonic2023convolutional} advances the latent state on the grid.
In the mesh architecture shown in Figure~\ref{fig:architecture}, it takes the 39 encoder channels and returns 16 latent channels $z$.
Its residual blocks are modulated by FiLM conditioning \citep{perez2018film} with the sonic and Alfv\'en Mach numbers of $\zeta$, computed as area-weighted averages over the fluid.
This formulation omits $\nu$, $\eta$, $\eta_b$ and $\mu_b$ from the trunk inputs, since these coefficients appear only in the dissipative part of \eqref{eq:mhd} and its interface conditions.

The mesh head returns to the fluid nodes and produces the potentials of the ideal increment.
Bilinear interpolation carries $z$ from the grid to a fluid node $v$ as $z_v$, and a pointwise network gives a latent velocity $\dot z_v$.
Three learned kernels $K_j$, $j=0,1,2$, mix the latent velocity in space.
Coefficients $a_j(z_v,f_v)$ from a second network combine the three outputs into six potentials,
\begin{equation}
p_v=\sum_{j=0}^{2}a_j(z_v,f_v)\,(K_j\dot z)_v,\qquad p=(\Phi_1,\Phi_2,\Pi_{11},\Pi_{12},\Pi_{22},\psi), \label{eq:potentials}
\end{equation}
a mass flux $\boldsymbol\Phi$, a symmetric momentum flux $\Pi$ and a scalar potential $\psi$ of the magnetic field.
The potentials are linear in the latent velocity $\dot z$ and nonlinear in $z$.
This form follows how a state changes when it is decoded from a latent variable \citep{lee2020model}.
If the state is a nonlinear function of $z$, its increment over a short time is the Jacobian of that function at $z$ applied to the latent velocity.
The coefficients $a_j(z)$ play the role of this state-dependent Jacobian, and $\dot z$ carries the direction and the size of the change.
A zero latent velocity then gives a zero increment, and the increment is tied to a rate of change rather than to the state itself.
\ref{app:mesh-head} defines the mesh kernels.

The conservative decoder $\mathcal C$ turns the potentials into increments of the state in weak form.
The density changes by the discrete divergence of the mass flux $\boldsymbol\Phi$, the momentum by the discrete divergence of the momentum flux $\Pi$, and the edge fluxes of the magnetic field by the discrete curl of $\psi$. On an oriented edge this curl is the value of $\psi$ at the end node minus its value at the start node.
Three properties then hold for any network weights.
The discrete divergence of a flux sums to zero over the mesh, which conserves the total mass.
The curl of a nodal potential has zero divergence on every triangle, which keeps the divergence of $\vB$ exact.
The momentum increment vanishes at the wall nodes, which keeps the no-slip condition.
\ref{app:decoder} gives the discrete updates.

\subsection{Learnable Dissipation}
\label{sec:learnable}

The dissipation is easier to compute than the ideal step.
However, stronger wall-normal magnetic fields or lower viscosity or magnetic diffusivity can thin the Hartmann layers. Resolving these layers on a finer mesh requires more Chebyshev terms in the exact half-step $\mathcal M_{\halfdt}$, which can add a considerable overhead relative to the cost of the trunk.
In this section, we replace the exact half-step with a learned half-step $\widehat{\mathcal M}_\phi$ with parameters $\phi$, whose cost per node is fixed.
We call this variant of the architecture \methodplus{}.

Diffusion mixes nearby values over a distance set by the diffusivity and the time step. In a homogeneous region away from boundaries, scalar diffusion is a convolution with a Gaussian heat kernel \citep{carslaw1959conduction}. Its width is set by the diffusion length $\ell$ \citep{jones2026constrained}, with $\ell=\sqrt{\nu\tau}$ for viscosity and $\ell=\sqrt{\eta\tau}$ for magnetic diffusion, where $\tau=\Delta t/2$.
The learned half-step groups nearby mesh nodes into clusters at the scale of $\ell$ and lets each node exchange information with clusters within a fixed multiple of that length.
For a fluid node $v$ away from the wall, the velocity increment is
\begin{equation}
\delta\vu_v=g_v^{(u)}\Big[\sum_{P\in\mathcal P_v}W_{vP}\,\big(\chi_1\vI+\chi_2\,\mathbf e_{vP}\mathbf e_{vP}^\top\big)(\bar\vu_P-\vu_v)-\lambda_v\vu_v\Big], \label{eq:student-visc}
\end{equation}
where $\mathcal P_v$ is the set of clusters paired with $v$, $\bar\vu_P$ is the area-weighted cluster velocity, and $\mathbf e_{vP}$ points from $v$ to the cluster center. The weights $W_{vP}$ set the strength of each exchange, $\chi_1$ and $\chi_2$ control its isotropic and directional parts, $\lambda_v$ sets the wall loss, and the learned factor $g_v^{(u)}$ scales the increment at $v$.

For magnetic diffusion, the exchange acts on the nodal stream function $\psi$ recovered from the edge fluxes after subtracting the uniform mean field. Oriented differences of the stream-function increment update the edge fluxes:
\begin{equation}
\delta\psi_v=g_v^{(\psi)}\sum_{P\in\mathcal P_v}\widetilde W_{vP}\,(\bar\psi_P-\psi_v),
\qquad (\delta b)_e=(G\,\delta\psi)_e=\delta\psi_{e^+}-\delta\psi_{e^-}, \label{eq:student-ohm}
\end{equation}
where $\bar\psi_P$ is the area-weighted cluster average and $e^-$ and $e^+$ are the start and end nodes of $e$. The oriented differences sum to zero around each triangle, so the increment preserves the discrete magnetic divergence. The weights $\widetilde W_{vP}$ also account for reflection and transmission across $\Gamma$ according to $\mu_b$.
\ref{app:learned} derives the viscous and Ohmic updates and their cluster weights.

The trained \method{} already contains layers that represent the ideal part of \eqref{eq:mhd}.
We therefore freeze $\mathcal I_\theta$ and use the exact half-step $\mathcal M_{\halfdt}$ of \method{} as the teacher for $\widehat{\mathcal M}_\phi$ in \methodplus{}, with the target $\mathcal M_{\halfdt}(\xi)$ for an input state $\xi$.
The input states are the states of the training trajectories, which enter the first half-step, and the states after the frozen ideal step, which enter the second half-step.
We also add training states perturbed by white noise in the velocity and in the magnetic field, because the trajectories carry little energy at the scale of the mesh, while in a rollout the half-step has to damp the small-scale errors of the ideal step.
We measure the relative error of the increment in the energy norm $\|\cdot\|_E$ \citep{morrison1980noncanonical}. The loss is
\begin{equation}
\mathcal J(\phi)=\mathbb E_\xi\,\frac{\big\|\widehat{\mathcal M}_\phi(\xi)-\mathcal M_{\halfdt}(\xi)\big\|_E^2}{\big\|\mathcal M_{\halfdt}(\xi)-\xi\big\|_E^2+\epsilon\,\|\xi\|_E^2}, \label{eq:distill}
\end{equation}
computed separately for the momentum and the magnetic field with $\epsilon=10^{-3}$.
The learned half-step never sees the next state of the data.

\section{Experimental Setup}
\label{sec:experimental-setup}

\subsection{Datasets}
\label{sec:datasets}

We simulate two datasets of freely decaying flows in 2D. Both datasets contain 51 snapshots per trajectory at intervals of $\Delta t=0.02$. The operating points vary $\nu$ and $\eta$ between $7.8\cdot10^{-4}$ and $4.9\cdot10^{-2}$, with initial values $M_s\in\{0.3,0.6,0.9\}$ and $M_A\in\{0.7,2\}$. The initial density is uniform and the velocity and magnetic fields are drawn from a divergence-free random-field generator. Solver settings and data quality checks are provided in \ref{app:experiment-configs}.

In the \textit{periodic box} dataset, the fluid fills the entire domain. We solve the problem with Dedalus \citep{burns2020dedalus} using a Fourier pseudospectral discretization on a $128\times128$ grid. We run several simulations per parameter combination with independently drawn initial conditions. We split the 3047 retained trajectories into 2757 training and 290 test trajectories.

In the \textit{permeable body} dataset, the flow encounters stationary permeable bodies. We solve the problem with NGSolve \citep{schoberl2014c} on body-fitted triangular meshes. For each combination of $(\nu,\eta,M_s,M_A)$, we use body permeabilities of $3$, $7.21$ and $26.8$ to span nearly an order of magnitude in permeability contrast. The bodies are smooth star-shaped curves with fixed total area. We measure shape complexity by $q$, the perimeter divided by that of a circle of the same area. A circle has $q=1$, and larger values indicate more complex shapes. The training and in-distribution (ID) test sets use a single body with $q\leq1.07$. One out-of-distribution (OOD) test set uses more complex shapes with $1.07<q\leq1.18$. The other uses two bodies with the same combined area as the single body in the ID set. We split the trajectories into 2006 for training, 144 for ID evaluation, 143 for the shape test and 138 for the topology test.

\begin{figure}[!htbp]
\centering
\includegraphics[width=0.92\linewidth]{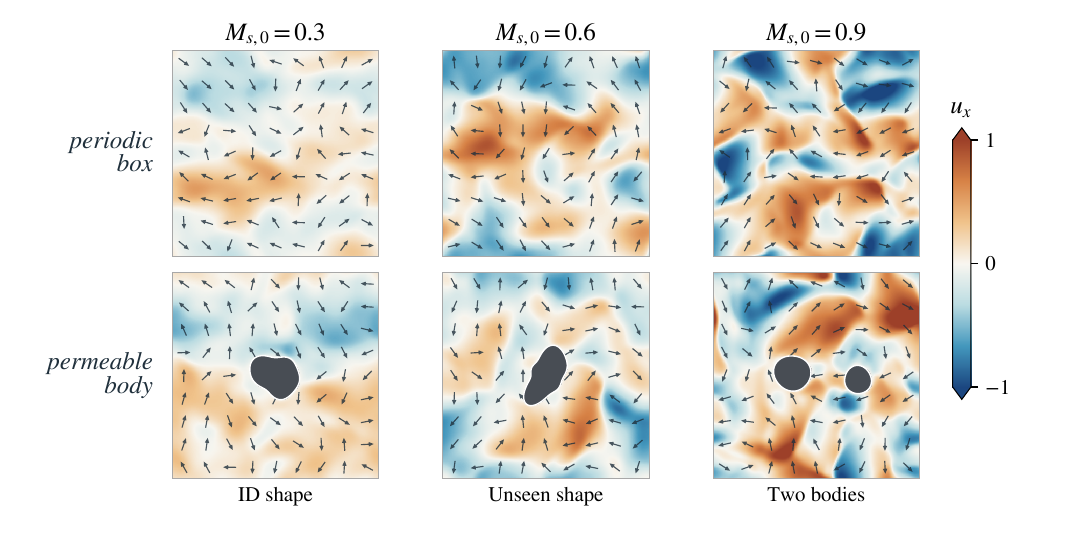}
\caption{Reference flows at $t=0.2$ in \textit{periodic box} (top) and \textit{permeable body} (bottom). Color shows the horizontal velocity $u_x$. Arrows give the direction of the magnetic field in the fluid, and gray regions mark the bodies. Each column pairs independent initial fields at the same viscosity, resistivity and initial Mach numbers, with $M_{A,0}=0.7$ throughout.}
\label{fig:dataset-samples}
\end{figure}

\begin{figure}[!htb]
\centering
\includegraphics[width=0.89\linewidth]{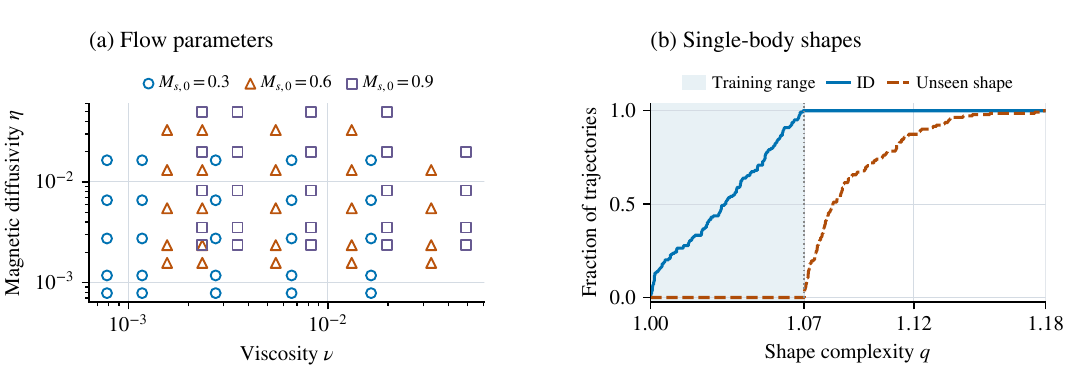}
\caption{Coverage of the reference datasets. (a) Viscosity and resistivity in the parameter sweep shared by \textit{periodic box} and \textit{permeable body}. Each symbol gives an initial sonic Mach number. Operating points that differ only in Alfv\'en Mach number coincide in this projection. (b) Cumulative distributions of shape complexity in the single-body ID and unseen-shape test sets of \textit{permeable body}. Shading marks the training range.}
\label{fig:dataset-coverage}
\end{figure}

Figure~\ref{fig:dataset-samples} shows one pair of reference flows at each initial sonic Mach number. The two trajectories in each column share an operating point. The lower row shows the single-body ID test and the two geometry shifts.
Figure~\ref{fig:dataset-coverage} shows which viscosity and resistivity pairs occur at each initial sonic Mach number. The distribution of $q$ also distinguishes two ways of testing a model on a new body. The ID test samples new shapes within the training range, while the shape OOD test lies beyond its upper limit.

\subsection{Baselines}
\label{sec:baselines}

On \textit{periodic box}, we compare energy attribution in \method{} with GENERIC-CNO \citep{sulskis2026generic} and tLaSDI \citep{park2024tlasdi}. GENERIC-CNO adapts the GENERIC formalism by using separate CNOs to represent the energy and entropy functionals. tLaSDI uses a skew-symmetric and a positive semidefinite operator on learned energy and entropy gradients, respectively.

For the rollout comparisons we use Transolver \citep{wu2024transolver} and MeshGraphNet \citep{pfaff2020learning} on both datasets. On \textit{periodic box} we use FNO \citep{li2020fourier} directly because the input and output points already lie on the Fourier grid. On \textit{permeable body}, GINO \citep{li2023geometry} adds graph operators that transfer fields between the body-fitted mesh and the Fourier grid. We include DoMINO \citep{ranade2025domino} on this dataset because its geometry encoder uses the distance to the obstacle and samples nearby points on its surface. These inputs are absent in the unobstructed domain. Each baseline predicts field increments at the grid points or mesh nodes for autoregressive rollout.

\subsection{Training Protocol}
\label{sec:training-protocol}

We use the same curriculum pushforward protocol \citep{brandstetter2022message} for \method{} and all baselines. We sample training windows of 10 consecutive frames. At each optimizer step, we draw a free rollout length uniformly between zero and the current limit. This limit rises from zero to eight over the first 6000 optimization steps. We execute the free rollout without gradients and differentiate the loss of the next prediction against the corresponding reference frame. Training uses a relative squared $L^2$ loss on standardized channels, a batch of 16 trajectories and gradient norm clipping at 1. We use a 500-step learning-rate warmup and a cosine schedule over 30\,000 optimization steps. On \textit{periodic box} we stop after 15\,000 updates for all models. We use AdamW for FNO and GINO and Adam for the other models. On \textit{periodic box}, the peak learning rate is $5\cdot10^{-4}$ for \method{}, $10^{-3}$ for FNO and Transolver, and $10^{-4}$ for MeshGraphNet. For the \textit{permeable body} baselines we take the peak rates from their published configurations, $10^{-4}$ for MeshGraphNet and $10^{-3}$ for Transolver, DoMINO and GINO.

For \methodplus{}, we freeze the ideal step from the 15\,000-step checkpoint on \textit{periodic box} and the 30\,000-step checkpoint on \textit{permeable body}. On each dataset we train only the learned dissipative half-step for 20\,000 updates with AdamW and a peak learning rate of $3\cdot10^{-4}$. On \textit{periodic box}, each batch contains 14 reference states, 14 states after the frozen ideal step and four reference states with high-frequency perturbations. On \textit{permeable body}, each batch contains seven reference states, seven states after the frozen ideal step and two reference states perturbed with noise. For these last two states, we add white noise to the velocity at free fluid nodes and take differences of a random nodal potential to perturb the magnetic edge fluxes while preserving their discrete divergence. For these mesh updates we use the loss in \eqref{eq:distill}.

\subsection{Evaluation}
\label{sec:evaluation}

Starting from the first reference snapshot, each model predicts the next 50 frames autoregressively. For each field $c$ and rollout step $k$, we compute the normalized root mean squared error
\[
\operatorname{NRMSE}_{c,k}
=\frac{1}{\sigma_c}\left(\frac{\sum_i w_i\big(\widehat{\xi}_{c,k}(x_i)-\xi_{c,k}(x_i)\big)^2}{\sum_i w_i}\right)^{1/2},
\]
where $\widehat{\xi}_{c,k}(x_i)$ and $\xi_{c,k}(x_i)$ are the predicted and reference values of field $c$ at spatial sample $x_i$, $w_i$ is its quadrature weight over the fluid, and $\sigma_c$ is the standard deviation of channel $c$ on a common training reference. The same scales are used across models and evaluation splits for each dataset. The fixed scale makes errors comparable across fields and prevents frames with small spatial fluctuations from dominating the average. We report the mean across test trajectories after $100\%$ of the rollout steps.

We also measure free-energy errors. The nondimensional isothermal free energy, with the fluid term from \citet{zhang2023global}, is
\begin{equation}
F(\xi)=
\int_{\Omega_f}\!\left[\tfrac12\rho|\vu|^2+
  \rho\log\rho-\rho+1+\tfrac12|\vB_f|^2\right]\,dx
+\int_{\Omega_b}\!\frac{|\vB_b|^2}{2\mu_b}\,dx .
\label{eq:eval-free-energy}
\end{equation}
Each probe starts from a reference frame $\xi$. We normalize the changes by the reference loss $\varepsilon=F(\xi)-F(\xi^+)$ to the next frame. For \method{}, $\zeta$ is the state after the first dissipative half-step, $y$ is the state after the ideal update, and $\widehat{\xi}$ is the full prediction. For the additive baselines, we set $\zeta=\xi$ and obtain $y$ by retaining only the reversible term before mapping back to the physical fields. The two energy ratios are
\begin{equation}
r_L^{\mathrm{seq}}=\frac{F(y)-F(\zeta)}{\varepsilon},
\qquad
r_M^{\mathrm{seq}}=\frac{F(\xi)-F(\zeta)+F(y)-F(\widehat{\xi})}{\varepsilon}.
\label{eq:eval-attribution}
\end{equation}
An energy-neutral ideal update gives $r_L^{\mathrm{seq}}=0$. The dissipative updates account for the reference loss when $r_M^{\mathrm{seq}}=1$. We also measure relative mass errors and discrete magnetic divergence.

We also test how well the ideal update preserves the magnetic-flux invariant $C_2=\tfrac12\int_{\Omega_f}\rho\tilde\psi^2\,dx$ \citep{andreussi2012hamiltonian}. Here $\tilde\psi$ is the stream function after removal of the mean magnetic field. On \textit{periodic box}, we drop the zero Fourier mode (the grid average) and solve a Poisson equation in Fourier space. On \textit{permeable body}, we take the area-weighted mean of the Raviart-Thomas field over the fluid and the body, subtract the fluxes of this uniform field and integrate the residual fluxes along a spanning tree. We fix the additive constant by subtracting the fluid mean of the potential, weighted by $\rho$ on the grid and by $\rho_v A_v$ on the mesh. The ideal equations advect the flux function with the fluid, and mass conservation preserves its mass-weighted second moment. We measure the absolute change in $C_2$ under the ideal update divided by its absolute change over the reference step.

We test extrapolation in $\mathrm{Re}$ and $\mathrm{Rm}$ on both datasets. We retrain models separately along each axis on operating points at initial levels $2$, $5$ and $12$. For each axis, evaluation uses the trajectories at levels $28$ and $42$ that were held out from the restricted training set. On \textit{periodic box} there are 1275 such trajectories for $\mathrm{Re}$ and 1274 for $\mathrm{Rm}$. On \textit{permeable body} there are 877 for $\mathrm{Re}$ and 868 for $\mathrm{Rm}$. To test geometric generalization, we use the full-training checkpoints on \textit{permeable body} for the shape and topology OOD tests.

\section{Experimental Results}
\label{sec:results}

\subsection{Energy Attribution Compared with Metriplectic Baselines}
\label{sec:results-periodic}

The metriplectic models in Section~\ref{sec:baselines} build a split between ideal and dissipative dynamics into the architecture, and hence \method{} is not the first attempt to employ the metriplectic formalism in a neural operator architecture. We begin with a comparison to methods based on this formalism. We pay particular attention to energy attribution because a model can predict a decrease in free energy over a full step even when its nominally reversible branch supplies that decrease. Thus, this check is necessary to understand whether the model correctly represents the reversible and irreversible parts of the system. Since GENERIC-CNO and tLaSDI operate on regular grids, we compare them with \method{} on \textit{periodic box} only. All models in this section follow the training protocol in Section~\ref{sec:training-protocol} and stop after 15{,}000 updates. Table~\ref{tab:periodic-rollout} compares rollout error, conservation diagnostics and energy attribution.

\begin{table}[!htbp]
\centering
\caption{Results on \textit{periodic box} test trajectories. At the final rollout step, we report mean NRMSE and magnetic divergence, together with the relative drifts in mass and free energy from their reference values. The attribution columns give medians over 725 pairs of consecutive reference frames, starting at frames 0, 10, 20, 30 and 40 of every other test trajectory.}
\label{tab:periodic-rollout}
\small
\setlength{\tabcolsep}{4.5pt}
\begin{tabular*}{\linewidth}{@{\extracolsep{\fill}}lrrrrrrr@{}}
\toprule
Model & NRMSE & Mass error & Free-energy error & Magnetic div. & $r_L^{\mathrm{seq}}$ & $r_M^{\mathrm{seq}}$ & $C_2$ drift \\
\midrule
\method{}   & 0.1056 & $9.54\times10^{-7}$ & $1.68\times10^{-2}$ & $4.43\times10^{-5}$ & $-0.019$ & $0.999$ & $0.056$ \\
tLaSDI      & 0.4095 & $1.29\times10^{-2}$ & $1.38\times10^{-1}$ & $1.68\times10^{0}$ & $-0.906$ & $0.096$ & $0.769$ \\
GENERIC-CNO & 1.8470 & $1.61\times10^{-1}$ & $7.24\times10^{0}$ & $7.28\times10^{0}$ & $-1.150$ & $-0.469$ & $2.087$ \\
\bottomrule
\end{tabular*}
\end{table}

For both baselines, the nominally reversible term removes at least $90\%$ of the reference energy loss at the median. Training on successive states constrains the combined update. Energy removed by the reversible term can compensate for missing dissipation in the other branch while improving the fit to the next frame. In \method{}, the ideal update changes the free energy and the magnetic-flux invariant by small fractions of their reference-step changes, and the dissipative half-steps account for nearly the full reference energy loss. Thus, of the three models, \method{} comes closest to the physical division between transport and dissipation on these probes.

\subsection{Validation on In-Distribution Data}
\label{sec:results-fem-id}

We first test how well \method{} interpolates within the training distribution. For \textit{periodic box}, the operating points are represented in training, but each test trajectory starts from an independently drawn flow field. Table~\ref{tab:periodic-id} compares the models on these trajectories at a forward cost of about $21$ GFLOP per step.

\begin{table}[!htbp]
\centering
\caption{In-distribution results on \textit{periodic box}. At step 50 we report mean NRMSE and magnetic divergence, and median absolute relative errors in mass and free energy, across 290 test trajectories. The attribution columns give medians over 725 pairs of consecutive reference frames, starting at frames 0, 10, 20, 30 and 40 of every other test trajectory. A dash denotes a branch diagnostic that a single-update model does not define.}
\label{tab:periodic-id}
\small
\setlength{\tabcolsep}{4.5pt}
\begin{tabular*}{\linewidth}{@{\extracolsep{\fill}}lrrrrrrr@{}}
\toprule
Model & NRMSE & Mass error & Free-energy error & Magnetic div. & $r_L^{\mathrm{seq}}$ & $r_M^{\mathrm{seq}}$ & $C_2$ drift \\
\midrule
\method{}    & 0.1056 & $9.54\times10^{-7}$ & $1.68\times10^{-2}$ & $4.43\times10^{-5}$ & $-0.019$ & $0.999$ & $0.056$ \\
FNO          & 0.2441 & $9.06\times10^{-3}$ & $6.91\times10^{-2}$ & $4.45\times10^{-1}$ & \tabledash & \tabledash & \tabledash \\
Transolver   & 0.4469 & $1.09\times10^{-2}$ & $2.31\times10^{-1}$ & $8.36\times10^{0}$ & \tabledash & \tabledash & \tabledash \\
MeshGraphNet & 0.6902 & $9.85\times10^{-2}$ & $3.28\times10^{-1}$ & $6.54\times10^{0}$ & \tabledash & \tabledash & \tabledash \\
\bottomrule
\end{tabular*}
\end{table}

\par\addvspace{\intextsep}
\noindent\begin{minipage}{\linewidth}
\makeatletter\def\@captype{figure}\makeatother
\centering
\includegraphics[width=0.82\linewidth]{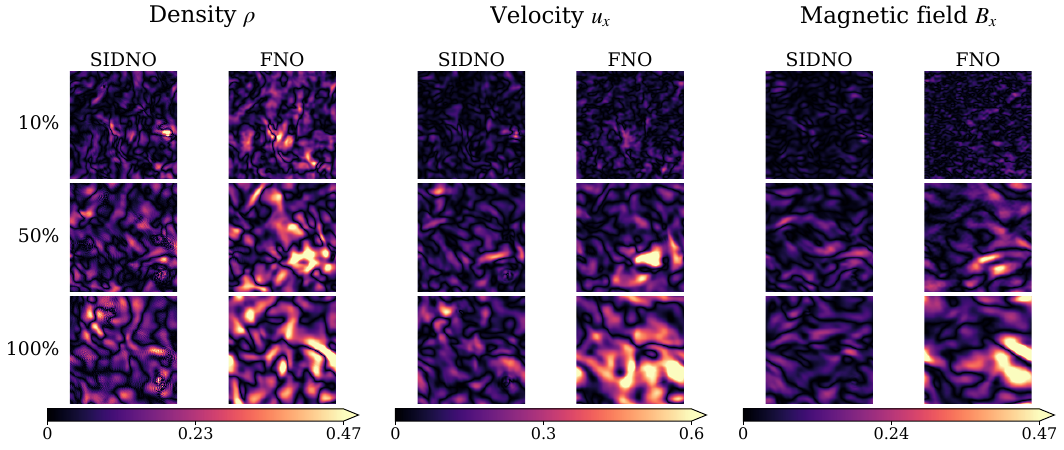}
\caption{Pointwise errors for \method{} and FNO on an ID test trajectory from \textit{periodic box} at steps 5, 25 and 50. The colors show $|\widehat{\xi}_c-\xi_c|/\sigma_c$. Each field shares a color scale across models and frames, with saturation at the pooled 99th percentile.}
\label{fig:periodic-id-fields}
\end{minipage}\par\addvspace{\intextsep}

\begin{wrapfigure}[12]{r}{0.42\textwidth}
\vspace{-\intextsep}
\setlength{\abovecaptionskip}{4pt}
\centering
\includegraphics[width=\linewidth]{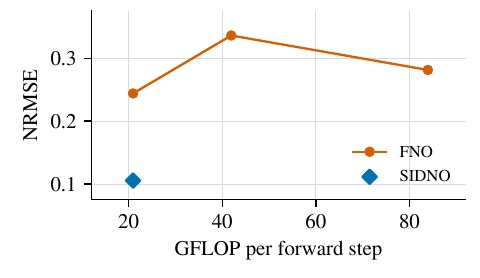}
\caption{Final-step ID NRMSE on \textit{periodic box} versus the cost of one forward step. FNO widths are 181, 258 and 367 (left to right, seed zero).}
\label{fig:periodic-scaling}
\end{wrapfigure}
After 50 steps on \textit{periodic box}, the mean NRMSE is $0.1056$ for \method{}, $0.2441$ for FNO, $0.4469$ for Transolver and $0.6902$ for MeshGraphNet. For \method{}, the median relative mass error remains below $10^{-6}$, while its median relative free-energy error reaches $1.68\%$. At the matched budget of about $21$ GFLOP per step, FNO gives the lowest error among the three baselines. Increasing the FNO width to 258 and 367 gives errors of $0.3362$ and $0.2814$ at $41.92$ and $83.93$ GFLOP per step, respectively (Figure~\ref{fig:periodic-scaling}). The initial FNO width of 181 is therefore sufficient to achieve the lowest error observed for this architecture on \textit{periodic box} under the chosen training schedule.

\par\WFclear

On \textit{permeable body}, the mesh encoder transfers the flow fields to the latent grid, and the mesh head maps the latent features back to the fluid nodes. We test whether the branches retain their physical roles through these transfers and compare rollout accuracy with that of operators designed for irregular meshes.
We fix the inference budget at approximately $24$ GFLOP per step and adjust the baseline widths to match it.

\begin{figure}[!htbp]
\centering
\includegraphics[width=0.94\linewidth]{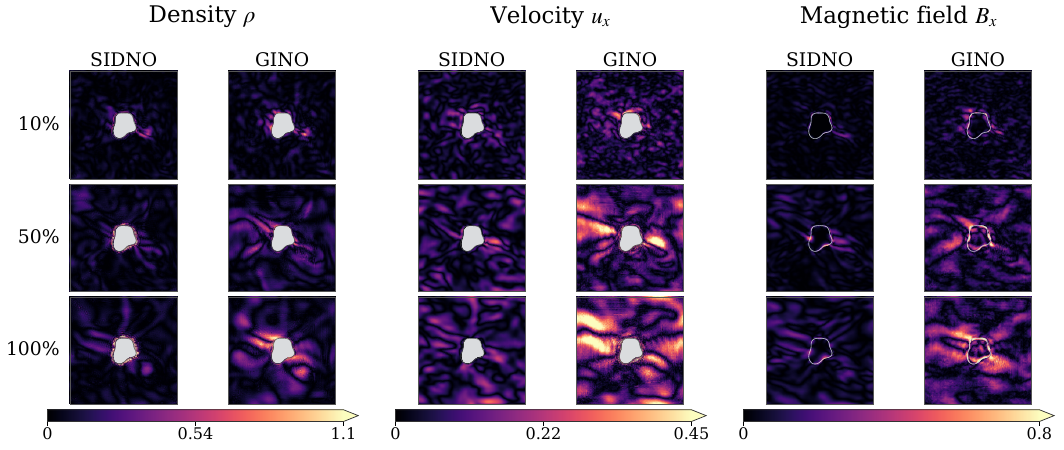}
\caption{Pointwise errors for \method{} and GINO on an ID test trajectory from \textit{permeable body}. The colors show $|\widehat{\xi}_c-\xi_c|/\sigma_c$. Each field shares a color scale across models and frames, with saturation at the pooled 99th percentile.}
\label{fig:fem-id-fields}
\end{figure}

\begin{table}[!htbp]
\centering
\caption{In-distribution results on \textit{permeable body} at comparable FLOP per forward step. At the final rollout step, we report mean NRMSE and magnetic divergence, together with the relative drifts in mass and free energy from their reference values. Attribution medians use five pairs of consecutive reference frames from each of the 144 test trajectories. The four baselines predict a single update per step, which precludes separate ideal and dissipative measurements.}
\label{tab:fem-id}
\small
\setlength{\tabcolsep}{4.5pt}
\begin{tabular*}{\linewidth}{@{\extracolsep{\fill}}lrrrrrrr@{}}
\toprule
Model & NRMSE & Mass error & Free-energy error & Magnetic div. & $r_L^{\mathrm{seq}}$ & $r_M^{\mathrm{seq}}$ & $C_2$ drift \\
\midrule
\method{}    & 0.0989 & $4.24\times10^{-9}$ & $2.50\times10^{-2}$ & $8.41\times10^{-5}$ & $-0.019$ & $1.011$ & $0.039$ \\
GINO         & 0.2187 & $3.32\times10^{-3}$ & $5.71\times10^{-2}$ & $1.00\times10^{0}$ & \tabledash & \tabledash & \tabledash \\
Transolver   & 0.5367 & $4.94\times10^{-3}$ & $3.58\times10^{-1}$ & $9.63\times10^{-1}$ & \tabledash & \tabledash & \tabledash \\
MeshGraphNet & 0.6115 & $2.32\times10^{-2}$ & $3.79\times10^{-1}$ & $1.00\times10^{0}$ & \tabledash & \tabledash & \tabledash \\
DoMINO       & 0.5987 & $5.58\times10^{-3}$ & $4.19\times10^{-1}$ & $9.86\times10^{-1}$ & \tabledash & \tabledash & \tabledash \\
\bottomrule
\end{tabular*}
\end{table}

Table~\ref{tab:fem-id} shows that at this budget, \method{} has the lowest rollout error.
On the \textit{permeable body} ID data, the median $r_L^{\mathrm{seq}}$ is $-0.019$ and its 90th percentile is $0.023$. The median $r_M^{\mathrm{seq}}$ is $1.011$ and its 90th percentile is $1.046$. The median $C_2$ drift under the ideal update is $0.039$.

\noindent\begin{minipage}{\linewidth}
\begin{wrapfigure}[12]{r}{0.42\textwidth}
\vspace{-\intextsep}
\setlength{\abovecaptionskip}{4pt}
\centering
\includegraphics[width=\linewidth]{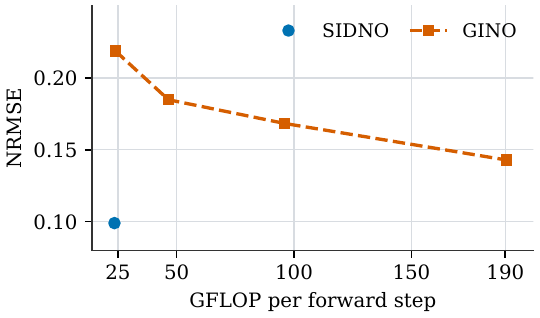}
\caption{ID NRMSE on \textit{permeable body} after $100\%$ of the rollout steps versus forward cost.}
\label{fig:fem-scaling}
\end{wrapfigure}
\noindent Even with the transfers between the mesh and the latent grid, the ideal branch removes only about $2\%$ of the reference energy loss at the median, as on \textit{periodic box}. The dissipative half-steps account for nearly all of that loss on both grids. The two branches of \method{} thus retain their distinct physical roles even on complex meshes.

GINO is the closest baseline at the matched budget. In Figure~\ref{fig:fem-scaling}, we increase its width to test whether additional FLOP closes the gap. Its NRMSE decreases from $0.2187$ to $0.1848$ at $46.5$ GFLOP and $0.1683$ at $96.0$ GFLOP. At $190.6$ GFLOP, about eight times the cost of one \method{} step, GINO reaches $0.1428$ on the same trajectories.

\par\WFclear
\end{minipage}\par

\subsection{Validation on Out-of-Distribution Data}
\label{sec:results-ood}

The ID tests vary the initial fields and problem parameters within the training distribution. At higher initial $\mathrm{Re}$ or $\mathrm{Rm}$, viscosity or magnetic diffusion becomes weaker relative to transport. We test whether a model trained at the lower levels can still predict the evolution of these more weakly damped flows.

\par\addvspace{\intextsep}
\noindent\begin{minipage}{\linewidth}
\makeatletter\def\@captype{figure}\makeatother
\centering
\includegraphics[width=0.81\linewidth]{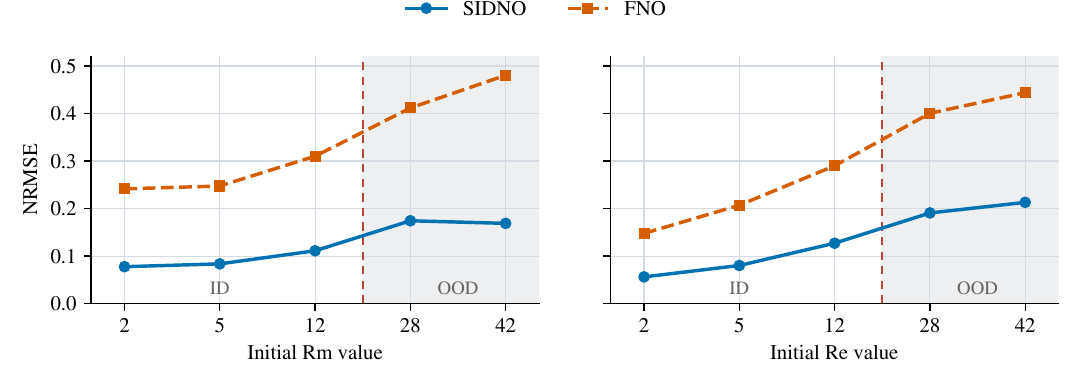}
\caption{Mean NRMSE on \textit{periodic box} after $100\%$ of the rollout steps versus the initial value of $\mathrm{Rm}$ (left) or $\mathrm{Re}$ (right). The red dashed line separates the ID values $2$, $5$ and $12$ from the shaded OOD values $28$ and $42$.}
\label{fig:periodic-rerm}
\end{minipage}\par\addvspace{\intextsep}

On \textit{periodic box}, each panel of Figure~\ref{fig:periodic-rerm} follows a separate \method{} model from the three levels used in training to the two held-out levels. For $\mathrm{Re}$, the mean NRMSE rises from $0.1271$ at level $12$ to $0.1908$ and $0.2131$ at levels $28$ and $42$. For $\mathrm{Rm}$, the error rises from $0.1112$ at level $12$ to $0.1743$ at level $28$ and stays close to that value at level $42$. Across all held-out trajectories, the mean error is $0.2024$ for $\mathrm{Re}$ and $0.1715$ for $\mathrm{Rm}$. For FNO, the corresponding errors are $0.4232$ and $0.4476$.

On \textit{permeable body}, we repeat the Reynolds-number tests and also change the geometry while keeping the full-training checkpoints fixed. The first geometry test uses more irregular obstacle shapes. The second changes the topology by adding a second body.
Table~\ref{tab:fem-ood} shows that \method{} has the lowest NRMSE on both geometry tests. A more irregular boundary raises its error only from $0.0989$ to $0.1067$. Adding a second obstacle nearly doubles the error to $0.1964$, still less than half the GINO error of $0.4539$ on those domains.

\par\addvspace{6pt}
\noindent\begin{minipage}{\linewidth}
\makeatletter\def\@captype{table}\makeatother
\setlength{\abovecaptionskip}{5pt}\setlength{\belowcaptionskip}{5pt}\centering
\caption{Mean NRMSE after $100\%$ of the rollout steps for the geometry shifts on \textit{permeable body}. We use the checkpoints from Table~\ref{tab:fem-id} for all three test sets.}
\label{tab:fem-ood}
\small
\setlength{\tabcolsep}{4.5pt}
\begin{tabular*}{\linewidth}{@{\extracolsep{\fill}}lrrr@{}}
\toprule
Model & ID & Unseen shapes & Two bodies \\
\midrule
\method{}    & 0.0989 & 0.1067 & 0.1964 \\
GINO         & 0.2187 & 0.2278 & 0.4539 \\
Transolver   & 0.5367 & 0.5342 & 0.5162 \\
MeshGraphNet & 0.6115 & 0.6088 & 0.5767 \\
DoMINO       & 0.5987 & 0.5944 & 0.5637 \\
\bottomrule
\end{tabular*}
\end{minipage}\par\addvspace{6pt}

\par\addvspace{\intextsep}
\noindent\begin{minipage}{\linewidth}
\makeatletter\def\@captype{figure}\makeatother
\centering
\includegraphics[width=0.81\linewidth]{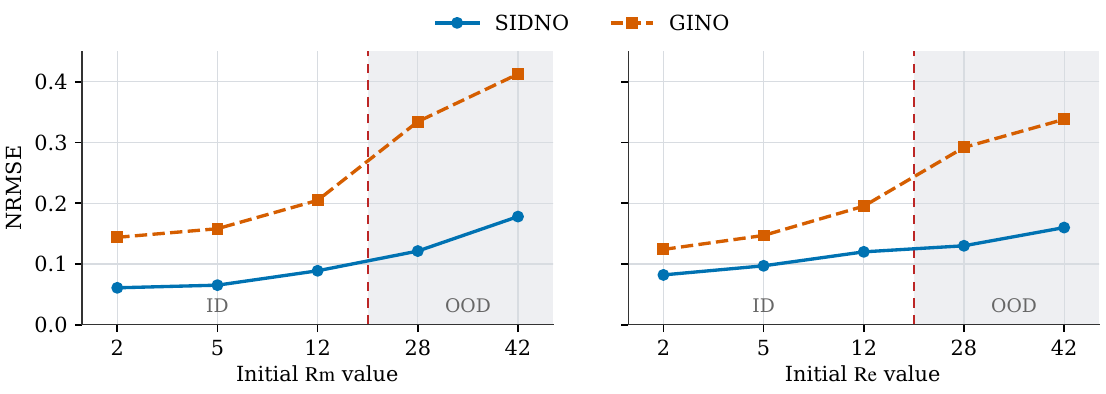}
\caption{Mean NRMSE on \textit{permeable body} after $100\%$ of the rollout steps versus the initial value of $\mathrm{Rm}$ (left) or $\mathrm{Re}$ (right). The red dashed line separates the ID values $2$, $5$ and $12$ from the shaded OOD values $28$ and $42$.}
\label{fig:fem-rm-levels}
\end{minipage}\par\addvspace{\intextsep}

Figure~\ref{fig:fem-rm-levels} demonstrates that in the $\mathrm{Rm}$ test, \method{} reaches $0.1492$ at the held-out levels, compared with $0.3730$ for GINO. At the held-out $\mathrm{Re}$ levels $28$ and $42$, \method{} has less than half the NRMSE of GINO. The advantage over the strongest ID baseline thus extends to both changes in operating conditions as well as to changes in geometry.

\subsection{Learnable Dissipation}
\label{sec:results-distillation}

A learned dissipative half-step can reduce the cost of the update on finely resolved meshes. We first test on both datasets how replacing the analytic half-step affects rollout accuracy and whether the energy loss remains attributable to dissipation. \methodplus{} uses the learned half-step with the same ideal branch as \method{}.

Table~\ref{tab:periodic-distillation} shows that on \textit{periodic box}, replacing the analytic half-step raises the final-step NRMSE from $0.1056$ to $0.1090$. The learned half-steps account for a median $99.1\%$ of the reference free-energy loss, compared with $99.9\%$ for the teacher. The ideal energy ratio and the magnetic-flux drift remain the same to the precision shown in the table.

\begin{table}[!htbp]
\centering
\caption{ID accuracy and energy attribution on \textit{periodic box} for the teacher and the distilled model. NRMSE is measured after $100\%$ of the rollout steps. Attribution medians use the same 725 pairs of consecutive reference frames as Table~\ref{tab:periodic-rollout}.}
\label{tab:periodic-distillation}
\small
\begin{tabular*}{\linewidth}{@{\extracolsep{\fill}}lrrrr@{}}
\toprule
Model & NRMSE & $r_L^{\mathrm{seq}}$ & $r_M^{\mathrm{seq}}$ & $C_2$ drift \\
\midrule
\method{}     & 0.1056 & $-0.019$ & 0.999 & 0.056 \\
\methodplus{} & 0.1090 & $-0.019$ & 0.991 & 0.056 \\
\bottomrule
\end{tabular*}
\end{table}

\begin{table}[!htbp]
\centering
\caption{ID accuracy and energy attribution on \textit{permeable body} for the teacher and the distilled model. NRMSE is measured after $100\%$ of the rollout steps. Attribution medians use five pairs of consecutive reference frames from each of the 144 test trajectories.}
\label{tab:distillation}
\small
\begin{tabular*}{\linewidth}{@{\extracolsep{\fill}}lrrrr@{}}
\toprule
Model & NRMSE & $r_L^{\mathrm{seq}}$ & $r_M^{\mathrm{seq}}$ & $C_2$ drift \\
\midrule
\method{}     & 0.0989 & $-0.019$ & 1.011 & 0.039 \\
\methodplus{} & 0.1385 & $-0.018$ & 0.929 & 0.038 \\
\bottomrule
\end{tabular*}
\end{table}

\noindent\begin{minipage}{\linewidth}
\begin{wrapfigure}{r}{0.39\textwidth}
\vspace{-\intextsep}
\centering
\includegraphics[width=\linewidth]{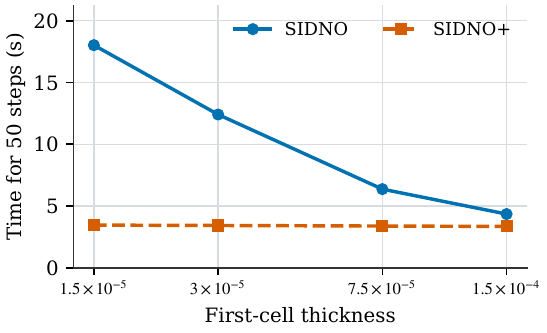}
\caption{Runtime for 50 prediction steps on \textit{permeable body}. The markers show the median over four cases at each first-cell thickness on an NVIDIA RTX A6000 with batch size one.}
\label{fig:distillation-timing}
\end{wrapfigure}
Table~\ref{tab:distillation} shows that distillation increases the ID rollout error on \textit{permeable body}. However, the learned half-steps still account for a median $92.9\%$ of the reference energy loss. The change in free energy and the normalized drift in the magnetic-flux invariant under the ideal update stay close to the teacher's values. The ID rollout error of \methodplus{} is still lower than that of all the baselines.

On \textit{permeable body}, the runtime benefit depends on the first-cell thickness next to the wall. To measure it, we refine that cell at four thicknesses. We time single-trajectory rollouts on an NVIDIA RTX A6000.
Figure~\ref{fig:distillation-timing} shows that reducing the first-cell thickness from $1.5\times10^{-4}$ to $1.5\times10^{-5}$ increases the median rollout time of \method{} from $4.35$ to $18.03$s. The median for \methodplus{} changes from $3.35$ to $3.46$s. At the smallest tested thickness, \methodplus{} completes the rollout in about one fifth of the teacher's time. As the near-wall mesh is refined, the larger time saving makes the switch from \method{} to \methodplus{} increasingly worthwhile.

\par\WFclear
\end{minipage}\par

\subsection{Ablation Study}
\label{sec:results-ablation}

On \textit{periodic box}, the physical fields already occupy the grid used by the trunk. Interpolation back to fluid nodes and the local kernels of the mesh head enter the update only on \textit{permeable body}. We therefore run the full sequence of ablations on its ID set, where we can measure the contribution of these mesh operations as well as that of the split. The first variant replaces the symmetric Strang composition with one full dissipative step before the ideal update. The second removes the analytic dissipation and keeps the conservative decoder. The third also replaces that decoder with an MLP that predicts nodal increments. The fourth removes the mesh head as well and predicts the increments directly from the interpolated latent features. Table~\ref{tab:fem-ablation} compares each variant with the full model.

\begin{table}[!htbp]
\centering
\caption{Ablation NRMSE on \textit{permeable body} after $100\%$ of the rollout steps.}
\label{tab:fem-ablation}
\small
\setlength{\tabcolsep}{5pt}
\begin{tabular*}{\linewidth}{@{\extracolsep{\fill}}p{0.43\linewidth}p{0.41\linewidth}r@{}}
\toprule
Variant & Change & NRMSE \\
\midrule
\method{} & Full model & 0.0989 \\
\method{} w/o Strang splitting & Apply one dissipative step before the ideal step & 0.1230 \\
\method{} w/o dissipation & apply only ideal branch & 0.3461 \\
\method{} w/o conservative decoder and splitting & Predict nodal increments with an MLP & 0.4119 \\
\method{} w/o mesh head, conservative decoder and splitting & Remove the mesh head, conservative decoder and analytic step & 0.5413 \\
\bottomrule
\end{tabular*}
\end{table}

Removing the analytic dissipative step raises NRMSE to $3.5$ times that of the full model even with the conservative decoder in place. Keeping dissipation and changing only its position in the step raises NRMSE by $24\%$. Without dissipation, the error rises from $0.3461$ to $0.4119$ when the conservative decoder is replaced and to $0.5413$ when the mesh head is also removed. The split into ideal and dissipative steps gives the largest reduction in error, and both the conservative decoder and the local mesh interactions improve the prediction even in its absence.

\section{Conclusion}
\label{sec:conclusion}

In this work, we proposed \method{}, which separates ideal and dissipative dynamics in compressible, isothermal MHD.
The operator uses Strang splitting to place a learned conservative ideal update between two analytic dissipative half-steps.
The finite-element decoder conserves mass and preserves the discrete magnetic divergence constraint.
On both \textit{periodic box} and \textit{permeable body}, the half-steps accounted for nearly all of the reference free-energy loss at the median.
Changes in free energy and the magnetic-flux invariant during the ideal update were small relative to their reference-step changes.
\method{} achieved lower final-frame NRMSE than both metriplectic baselines on \textit{periodic box}. On that dataset, its error increased at held-out Reynolds numbers, with a larger average error for the shift in $\mathrm{Re}$ than for the shift in $\mathrm{Rm}$. On \textit{permeable body}, it achieved the lowest error in the matched-cost ID comparison and retained its advantage under the tested geometry and Reynolds-number shifts.
After distillation, the learned half-steps in \methodplus{} still accounted for most of the reference free-energy loss on both datasets.

\appendix
\counterwithin*{table}{section}
\counterwithin*{figure}{section}

\section{Architecture Details}
\label{app:architecture}

\subsection{Dissipative Half-Step}
\label{app:dissipative}
At frozen $\rho$, the dissipative part of \eqref{eq:mhd} is the linear system
\begin{equation}
\left\{
\begin{aligned}
&\partial_t\vM=\nabla\cdot\big(2\nu\rho\,S(\nicefrac{\vM}{\rho})\big) && \text{in }\Omega_f,\\
&\partial_t\vB=\eta\,\nabla^2\vB && \text{in }\Omega_f,\\
&\partial_t\vB=\eta_b\,\nabla^2\vB && \text{in }\Omega_b,
\end{aligned}
\right.
\label{eq:dissipative}
\end{equation}
with the no-slip and interface conditions of Section~\ref{sec:problem}.
We discretize \eqref{eq:dissipative} with finite elements on the mesh of the trajectory. The velocity $\vu_v=\vM_v/\rho_v$ is given at the fluid vertices and interpolated linearly on each triangle. We use a lumped mass matrix $m$ and the Galerkin form of the viscous stress \citep{guermond2021second}. On the free fluid vertices the velocity satisfies
\[
m\,\dot\vu=-K_\rho\vu,\qquad
m=\operatorname{diag}_v(\rho_v A_v)\otimes I_2,\qquad
A_v=\sum_{\substack{T\subset\Omega_f\\T\ni v}}\frac{|T|}{3},
\]
where $|T|$ is the area of triangle $T$ and $I_2$ is the identity on the two velocity components. For nodal vectors $\vu,\mathbf w$ with piecewise linear interpolants $\vu_h,\mathbf w_h$ that vanish on the wall, the viscous stiffness matrix is given by
\[
\mathbf w^\top K_\rho\vu
=\sum_{T\subset\Omega_f}2\nu\bar\rho_T\,|T|\,
  \big[\operatorname{Tr}\!\left(S(\vu_h)^\top\nabla\mathbf w_h\right)\big]_T,\qquad
\bar\rho_T=\frac{1}{3}\sum_{v\in T}\rho_v .
\]

The magnetic field uses the lowest-order Raviart-Thomas space \citep{raviart1977mixed}. We write $\vB_h=\sum_e b_e\mathbf w_e$, where $b_e=\int_e\vB_h\cdot\vn_e\,ds$ is the signed normal flux and $\mathbf w_e$ has unit flux on edge $e$ and zero flux on the other edges. In two dimensions these fluxes couple to the out-of-plane electric field, a continuous piecewise linear scalar with nodal values $\boldsymbol E$ \citep{laakmann2022augmented}. The magnetic system and its edge-flux evolution are
\[
\dot b=-G\boldsymbol E,\qquad L\boldsymbol E=G^\top M_\mu b,\qquad
\dot b=-G L^{-1}G^\top M_\mu b .
\]
For an edge oriented from $e^-$ to $e^+$, take $\vn_e$ to be its right normal. The nodal difference matrix $G$, the magnetic mass matrix $M_\mu$ and the lumped electric mass matrix $L$ are
\[
G_{ev}=\delta_{v,e^+}-\delta_{v,e^-},\qquad
(M_\mu)_{ee'}=\sum_T\int_T\frac{\mathbf w_e\cdot\mathbf w_{e'}}{\mu_T}\,dx,\qquad
L_{vw}=\delta_{vw}\sum_{T\ni v}\frac{|T|}{3\widehat\eta_T}.
\]
Here $\delta$ is the Kronecker delta. The magnetic and electric mass matrices include triangles in both regions. The permeability and resistivity $(\mu_T,\widehat\eta_T)$ are $(1,\eta)$ in the fluid and $(\mu_b,\mu_b\eta_b)$ in the body.

Both systems have fixed matrices over the half-step $\tau=\halfdt$. With
\[
S_V=m^{-1/2}K_\rho m^{-1/2},\qquad
S_E=L^{-1/2}G^\top M_\mu G L^{-1/2},
\]
the exact semi-discrete updates have the exponential time-differencing form \citep{cox2002exponential}
\begin{equation}
\begin{aligned}
\vu(\tau)&=\vu(0)-\tau m^{-1/2}\varphi_1(-\tau S_V)\,m^{-1/2}K_\rho\,\vu(0),\\
b(\tau)&=b(0)-\tau G L^{-1/2}\varphi_1(-\tau S_E)\,L^{-1/2}G^\top M_\mu\,b(0).
\end{aligned}
\label{eq:etd}
\end{equation}
Here $\varphi_1(z)=(e^z-1)/z$ for $z\ne 0$ and $\varphi_1(0)=1$. The momentum after the half-step is $\vM_v(\tau)=\rho_v\vu_v(\tau)$.
We evaluate the actions of $e^{-\tau S_V}$ for velocity and $\varphi_1(-\tau S_E)$ for magnetic flux with truncated Chebyshev series, using one sparse matrix-vector product per recurrence step \citep{hochbruck2010exponential}.

\subsection{Encoder}
\label{app:encoder}
The encoder forms its input fields at fluid vertices. The velocity is $\vu_v=\vM_v/\rho_v$. For the magnetic field we first reconstruct the Raviart-Thomas field from the three edge fluxes on each fluid triangle \citep{raviart1977mixed}. If $\overline{\vB}_T$ denotes its mean over $T$, the nodal field is the area-weighted average
\[
\vB_{f,v}=\frac{\sum_{T\ni v,\,T\subset\Omega_f}|T|\,\overline{\vB}_T}
                 {\sum_{T\ni v,\,T\subset\Omega_f}|T|}.
\]
At a wall vertex this average uses the triangles on the fluid side. Together with density and velocity it gives the five channels $(\rho,u_1,u_2,B_1,B_2)$ of $\zeta_v$. We subtract the training mean of each channel and divide by its training standard deviation. Both statistics use the fluid dual areas $A_v$ as weights.

The first five grid channels interpolate these normalized fields using the barycentric coordinates of each query in its triangle. For a grid point inside the body, we project the query onto the nearest boundary edge and interpolate between the two end nodes. The signed distance and the body mask describe the position of the original grid point relative to $\Gamma$.

The learned part of the encoder uses the same query positions, including the projected positions for points inside the body. The lift $W$ maps the five normalized fields and the five node features to 32 channels. The node features are $\log(1+d_v/h_v)$, the two components of the wall normal, and $\cos(2\alpha_v)$ and $\sin(2\alpha_v)$. Here $d_v$ is the wall distance, $h_v$ is the short semiaxis of the local mesh ellipse, and $\alpha_v$ gives the direction of its long axis. The kernel network $\kappa_\theta$ has two hidden layers of width 32. For each neighboring fluid node, the kernel takes its offset divided by the disk radius and its five node features, then weights the lifted channels componentwise as in \eqref{eq:encoder}. We sum these contributions with the normalized dual-area weights $\omega_{yv}$. A disk with no fluid nodes contributes zero to all 32 kernel channels.

\subsection{Mesh Head}
\label{app:mesh-head}
Bilinear interpolation gives the grid value $z_v=(Rz)(x_v)$ at every fluid node. A pointwise multilayer perceptron combines $z_v$ with the local geometry features $f_v$ to give the latent velocity $\dot z_v=\mathrm{MLP}(z_v,f_v)$. Each kernel $K_j$ gathers latent velocities from a neighborhood of the target node $q$. The three levels $j=0,1,2$ use dilations $1$, $2$ and $4$.

The shape of each neighborhood follows the local mesh. Let $d_e$ be the displacement along a fluid edge incident to $q$. The matrix $\Sigma_q=2\,\mathrm{mean}_e\,d_e d_e^\top$ records both the lengths and the directions of these edges. At level $j$ we include fluid nodes $p$ whose displacement $d=x_p-x_q$ satisfies $d^\top\Sigma_q^{-1}d\le 4^j$. The resulting ellipse extends farther along the long edges of a wall-layer mesh than across its thin cells. We express the offset of each selected neighbor in stencil coordinates as $s_{qp}=2^{-j}\Sigma_q^{-1/2}d$.

In these coordinates the same nine positions $(a,b)\in\{-1,0,1\}^2$ are used at every target node. A neighbor distributes its latent velocity to at most four adjacent positions with bilinear weights
$\beta_{ab}(s)=\max(0,1-|s_1-a|)\max(0,1-|s_2-b|)$. A neighbor halfway between two positions on a stencil axis contributes half to each. The factor $\omega_{qp}$ weights each contribution by the area represented by node $p$. Applying a learned matrix $W^j_{ab}$ at each position gives
\begin{equation}
(K_j\dot z)_q=\sum_{a,b}W^j_{ab}\sum_p\omega_{qp}\,\beta_{ab}(s_{qp})\,\dot z_p,\qquad \omega_{qp}=\frac{9A_p}{\sum_{p'}A_{p'}}, \label{eq:kernels}
\end{equation}
where both sums over neighbors run over level $j$. On a uniform grid, where $\Sigma_q=h^2\vI$, the neighbors fall on the taps and $K_j$ is an ordinary dilated convolution. The coefficients in \eqref{eq:potentials} combine the three kernel outputs into the six nodal potentials.

\subsection{Conservative Decoder}
\label{app:decoder}
For each fluid vertex $v$, the $P_1$ hat function $\phi_v$ is the continuous, piecewise affine function equal to one at $v$ and zero at every other vertex \citep{schoberl2014c}. On each incident triangle it is the barycentric coordinate associated with $v$. Let $|T|$ denote the area of a fluid triangle $T$, and let $\boldsymbol\Phi_T$ and $\Pi_T$ be the averages of the corresponding potentials at its three vertices. We write $h=1/128$ for the distance between adjacent points along either axis of the periodic $128\times128$ latent grid on the unit square. The decoder reads
\begin{equation}
\Delta\rho_v=\frac{h}{A_v}\sum_{T\ni v}|T|\,\boldsymbol\Phi_T\cdot\nabla\phi_v,\qquad
\Delta\vM_v=\frac{h}{A_v}\sum_{T\ni v}|T|\,\Pi_T\nabla\phi_v,\qquad
\Delta b_e=h\,(\psi_{e^+}-\psi_{e^-}), \label{eq:decoder}
\end{equation}
with $\Delta\vM_v=0$ at the nodes on $\Gamma$, $\psi=0$ on $\Gamma$ and in the body, and $e^\pm$ the end nodes of the edge $e$.
These are the weak forms of $-\nabla\cdot\boldsymbol\Phi$, $-\nabla\cdot\Pi$ and the two-dimensional curl of a scalar, $\operatorname{curl}\psi=(\partial_2\psi,-\partial_1\psi)$. For an in-plane vector $\mathbf a=(a_1,a_2)$, we use $\operatorname{curl}\mathbf a=\partial_1a_2-\partial_2a_1$.
The hat functions sum to one, which gives $\sum_vA_v\,\Delta\rho_v=0$.
Near the wall, $\boldsymbol\Phi$ and $\psi$ are multiplied by $\min(1,\nicefrac{d}{h})$, with $d$ the distance to $\Gamma$, which keeps the thin first cells of the mesh from amplifying the increments.
A floor $\rho\ge0.05$, applied with the total mass conserved, completes the ideal step.

\section{Learned Half-Step}
\label{app:learned}

Each field has its own diffusion length $\ell$ of Section~\ref{sec:learnable}.
The nodes are binned by a square grid of step $\nicefrac{\ell}{2}$, separately in the fluid and in the body.
Each nonempty bin forms a cluster $P$ with the area $A_P=\sum_{v\in P}A_v$, the area-weighted centroid $x_P$ and the area means $\bar f_P=\nicefrac{\sum_{v\in P}A_vf_v}{A_P}$ of nodal values.
The wall nodes enter the velocity clusters with $\vu=0$.
A node is paired with the clusters within $4\ell$, and a velocity pair is kept only if the segment between the node and the cluster does not cross the body.
The tables depend on the mesh and on $\ell$ and are built without iterations.
The sample vector $\mathbf z$ contains the standardized values of $\log_{10}\nu$, $\log_{10}\eta$ and $\log_{10}\mu_b$, followed by $\log(\rho_{\max}/\rho_{\min})$ over the fluid nodes. The radial profiles and nodal gates use $\mathbf z$ to set the FiLM scales and shifts shared by all nodes of that sample.

\subsection{Derivation of the Viscous Update}
\label{app:viscous-derivation}
The ideal step advances the continuity equation, leaving $\rho$ fixed in time during the dissipative half-step $\tau=\halfdt$. Consider a smooth velocity field in the bulk with spatially constant $\rho$ and $\nu$. Summing over the repeated index $j$, the viscous stress in \eqref{eq:mhd-fluid} gives
\[
\begin{aligned}
 \partial_t u_i=(\mathcal L\vu)_i
 &=\rho^{-1}\partial_j(2\nu\rho S_{ij})
  =\nu\,\partial_j\!\left(\partial_j u_i+\partial_i u_j
                         -\tfrac23\delta_{ij}\nabla\cdot\vu\right)\\
 &=\nu\left[\nabla^2 u_i+\tfrac13\partial_i(\nabla\cdot\vu)\right].
\end{aligned}
\]
Temporarily omit the gradient of the divergence. Each velocity component $f$ then solves $\partial_t f=\nu\nabla^2 f$. Write $f(x)$ and $f^+(x)$ for its values at $t$ and $t+\tau$. On the plane the heat kernel has diffusion length $\ell=\sqrt{\nu\tau}$ \citep{carslaw1959conduction}. Its convolution with $f$ gives
\[
 \Gamma_\ell(\mathbf r)=\frac{1}{4\pi\ell^2}
                \exp\!\left(-\frac{|\mathbf r|^2}{4\ell^2}\right),\qquad
 f^+(x)=\int_{\mathbb R^2}\Gamma_\ell(\mathbf r)f(x+\mathbf r)\,d\mathbf r .
\]
Since the kernel has unit mass, subtracting the value at time $t$ gives
\[
\begin{aligned}
 \delta f(x)=f^+(x)-f(x)
 &=\int_{\mathbb R^2}\Gamma_\ell(\mathbf r) f(x+\mathbf r)\,d\mathbf r
   -f(x)\int_{\mathbb R^2}\Gamma_\ell(\mathbf r)\,d\mathbf r\\
 &=\int_{\mathbb R^2}\Gamma_\ell(\mathbf r)\big[f(x+\mathbf r)-f(x)\big]\,d\mathbf r .
\end{aligned}
\]

For velocity, let $\mathbf d(\mathbf r)=\vu(x+\mathbf r)-\vu(x)$ and $\mathbf e=\mathbf r/|\mathbf r|$. Averaging $\mathbf d$ component by component accounts for the Laplacian. To recover $\partial_1(\nabla\cdot\vu)$, the first component must also sample $\partial_{12}u_2$. Projecting $\mathbf d$ onto the line of the pair couples the two components through $(\mathbf e\mathbf e^\top\mathbf d)_1=e_1^2d_1+e_1e_2d_2$. Reflection about this line preserves the parallel component and reverses the perpendicular one. A linear exchange respecting this reflection weights the two components separately. We use the same radial Gaussian in both directions, with coefficient $a$ across the pair and $a+b$ along it,
\begin{equation}
\begin{aligned}
 \delta\vu(x)&=\int_{\mathbb R^2}\Gamma_\ell(\mathbf r)
 \big[a(\vI-\mathbf e\mathbf e^\top)+(a+b)\mathbf e\mathbf e^\top\big]\mathbf d(\mathbf r)\,d\mathbf r\\
 &=\int_{\mathbb R^2}\Gamma_\ell(\mathbf r)
 \big(a\vI+b\,\mathbf e\mathbf e^\top\big)\mathbf d(\mathbf r)\,d\mathbf r .
\end{aligned}
\label{eq:heat}
\end{equation}

The matrix in \eqref{eq:heat} is even in $\mathbf r$. Only the even part of the Taylor expansion contributes,
\[
 \frac{d_j(\mathbf r)+d_j(-\mathbf r)}{2}
 =\tfrac12 r_1^2\partial_{11}u_j+r_1r_2\partial_{12}u_j
  +\tfrac12 r_2^2\partial_{22}u_j+O(|\mathbf r|^4).
\]
Under $\Gamma_\ell$, the radius $R=|\mathbf r|$ is independent of the uniform angle of $\mathbf e$. The required moments are
\[
 \mathbb E[r_i r_j]=2\ell^2\delta_{ij},\qquad \mathbb E[R^2]=4\ell^2,\qquad
 \mathbb E[e_1^4]=\tfrac38,\qquad \mathbb E[e_1^2e_2^2]=\tfrac18 .
\]
The mixed term averages to zero in $\mathbf d$. In the projection it carries the factor $e_1e_2r_1r_2=R^2e_1^2e_2^2$. Substitution gives
\[
\begin{aligned}
 \mathbb E[\mathbf d]&=\ell^2\nabla^2\vu+O(\ell^4),\\
 \mathbb E[(\mathbf e\mathbf e^\top\mathbf d)_1]
 &=\tfrac{\ell^2}{4}\big(3\partial_{11}u_1+\partial_{22}u_1+2\partial_{12}u_2\big)+O(\ell^4)\\
 &=\tfrac{\ell^2}{4}\nabla^2u_1+\tfrac{\ell^2}{2}\partial_1(\nabla\cdot\vu)+O(\ell^4).
\end{aligned}
\]
Exchanging coordinate indices gives the second component. Combining the two contributions in \eqref{eq:heat} yields
\begin{equation}
 \delta\vu=\ell^2\Big[\big(a+\tfrac b4\big)\nabla^2\vu
              +\tfrac b2\nabla(\nabla\cdot\vu)\Big]+O(\ell^4).
\label{eq:generator}
\end{equation}
With $\ell^2=\nu\tau$, matching the bulk viscous increment $\tau\mathcal L\vu+O(\tau^2)$ fixes both coefficients,
\[
 \tfrac b2=\tfrac13,\qquad a+\tfrac b4=1
 \qquad\Longrightarrow\qquad b=\tfrac23,\quad a=\tfrac56 .
\]
The values $(a,b)$ that match the bulk half-step to first order in $\tau$ initialize the mesh coefficients $(\chi_1,\chi_2)$ in \eqref{eq:student-visc}.

\subsection{Viscous Weights}
At fluid node $v$, let $\mathbf s_P=x_P-x_v$ be the displacement to the centroid of cluster $P$ and $r_{vP}=|\mathbf s_P|$. The cluster area $A_P$ replaces the area element in \eqref{eq:heat}. The radial profile $m_\phi$ tapers to zero at $r=4\ell$ and modifies the raw cluster weight,
\[
 k_P(r)=A_P\exp\!\left(-\frac{r^2}{4\ell^2}+m_\phi(r/\ell,\mathbf z)\right),\qquad
 |m_\phi|\le5 .
\]
At a bulk node $v$ with wall distance $d_v>4\ell$, set $\lambda_v=0$. For this fixed node, define $w_P$ as the fraction of the total raw weight assigned to cluster $P$,
\[
 w_P=\frac{k_P(r_{vP})}{\sum_{Q\in\mathcal P_v}k_Q(r_{vQ})},\qquad
 \sum_{P\in\mathcal P_v}w_P=1 .
\]
A linear field has zero scalar diffusion, which on the mesh requires $\sum_P\omega_{vP}\mathbf s_P=0$. We impose this condition and rescale toward the Gaussian radial second moment $4\ell^2$ using
\[
\begin{aligned}
 \left(\sum_P w_P\mathbf s_P\mathbf s_P^\top\right)\mathbf c_v
     &=-\sum_P w_P\mathbf s_P, &
 w'_P&=w_P(1+\mathbf c_v\cdot\mathbf s_P),\\
 S_2&=\sum_P w'_P r_{vP}^2, &
 \omega_{vP}&=w'_P
 \begin{cases}
  \min(3,4\ell^2/S_2),&S_2>0,\\
  1,&S_2\le0.
 \end{cases}
\end{aligned}
\]
If the moment matrix is ill conditioned, we retain the normalized cluster weight, $\omega_{vP}=w_P$. The angular moments on the mesh depend on the directions of the centroids.

Within $4\ell$ of a no-slip wall, we approximate the boundary by the tangent line at the wall point nearest to $v$. Let $s$ be the coordinate along this line and $\varsigma$ the signed normal distance, positive in the fluid. The fluid in this local model occupies the half-plane $H=\{(s,\varsigma)\in\mathbb R^2:\varsigma>0\}$. Reflecting a source $y=(s,\varsigma)$ across the line to $y^*=(s,-\varsigma)$ gives $K_D(x,y)=\Gamma_\ell(x-y)-\Gamma_\ell(x-y^*)$. The missing kernel mass damps the local value,
\[
 \delta f(x)=\int_H K_D(x,y)\big[f(y)-f(x)\big]\,dy-[1-m(x)]f(x),
 \qquad m(x)=\int_H K_D(x,y)\,dy .
\]
Let $d_P$ be the mean wall distance in $P$. Normalizing by the sum of the direct and reflected weights retains the missing mass as the wall loss $\lambda_v$,
\begin{equation}
\begin{aligned}
 r^*_{vP}&=\sqrt{r_{vP}^2+4d_vd_P},\\
 \omega_{vP}&=\frac{k_P(r_{vP})-k_P(r^*_{vP})}
  {\sum_{Q\in\mathcal P_v}\big[k_Q(r_{vQ})+k_Q(r^*_{vQ})\big]},
 &\lambda_v&=1-\sum_{P\in\mathcal P_v}\omega_{vP}.
\end{aligned}
\label{eq:image}
\end{equation}

Let $h_v$ be the shortest edge incident to $v$. The direction $\mathbf e_{vP}=\mathbf s_P/r_{vP}$ is set to zero at coincident locations. The first three arguments of the gate $G_\phi$ describe node $v$. Its last argument $\mathbf z$ supplies the FiLM conditioning for the whole sample. At fixed density the velocity increment $\delta\vu_v$ gives a momentum increment $\rho_v\delta\vu_v$, where
\[
\begin{aligned}
 W_{vP}&=\omega_{vP}\frac{\rho_v+\bar\rho_P}{2\rho_v},\qquad
 (\chi_1,\chi_2)=(a,b)+\Delta\boldsymbol\chi_\phi(\mathbf z),\\
 g_v^{(u)}&=\operatorname{softplus}\!\left[G_\phi\!\left(
   \log\rho_v,\min(d_v/\ell,4),\operatorname{clip}(\log(h_v/\ell),-6,3),\mathbf z\right)\right],\\
 \delta\vu_v&=g_v^{(u)}\left[\sum_{P\in\mathcal P_v} W_{vP}
   (\chi_1\vI+\chi_2\,\mathbf e_{vP}\mathbf e_{vP}^\top)(\bar\vu_P-\vu_v)-\lambda_v\vu_v\right].
\end{aligned}
\]
We use $-6$ and $3$ as fixed hyperparameters to bound the mesh-resolution input to the gate. For the viscous diffusion length $\ell=\sqrt{\nu\tau}$, these limits let the gate distinguish edge lengths from $e^{-6}\ell\approx\ell/403$ to $e^3\ell\approx20\ell$. The interval gives more logarithmic range to edges finer than $\ell$. Beyond either endpoint, further refinement or coarsening gives the same input.

Distillation trains the radial profile $m_\phi$, the affine map $\Delta\boldsymbol\chi_\phi$ that shifts the coefficients of the full and projected velocity differences, and the gate network $G_\phi$ that scales the nodal increment. The moment, wall and density formulas determine $W_{vP}$ and $\lambda_v$ at each training step. We initialize the three components with
\[
 m_\phi\equiv0,\qquad \Delta\boldsymbol\chi_\phi\equiv\mathbf0,\qquad
 (\chi_1,\chi_2)=(\tfrac56,\tfrac23),\qquad g_v^{(u)}=1 .
\]
At these values the student applies the Gaussian exchange in \eqref{eq:heat} with the coefficients derived in \eqref{eq:generator} and the wall damping in \eqref{eq:image}. It starts from this discrete approximation of the viscous half-step and learns corrections from the teacher's increments on \textit{permeable body}.

\subsection{Derivation of the Ohmic Update}
The edge-flux update in \eqref{eq:student-ohm} uses differences of a scalar potential. To derive its exchange kernel, we approximate the fluid-body interface locally by its tangent line. Let $\varsigma$ be signed distance to this line, positive in the fluid. In this local model $\Gamma=\{\varsigma=0\}$ divides $\mathbb R^2$ into the fluid half-plane $H_f=\{\varsigma>0\}$ and the body half-plane $H_b=\{\varsigma<0\}$. Both materials have diffusivity $\eta$, with permeabilities $\mu_f=1$ and $\mu_b>0$. Write $\gamma_i=1/\mu_i$. For the unit normal $\vn=(n_1,n_2)$ into the fluid and the tangent $\mathbf t=(-n_2,n_1)$, the full potential $\Psi_i$ gives
\[
 \vB_i=\operatorname{curl}\Psi_i=(\partial_2\Psi_i,-\partial_1\Psi_i),\qquad
 \vB_i\cdot\vn=\partial_{\mathbf t}\Psi_i,\qquad
 \vB_i\cdot\mathbf t=-\partial_n\Psi_i .
\]
Writing $[q]=q_f-q_b$ for a jump on $\Gamma$, \eqref{eq:mhd-interface} gives $[\partial_{\mathbf t}\Psi]=0$ and $[\gamma\partial_n\Psi]=0$. The first condition makes $[\Psi]$ constant along $\Gamma$. We choose the additive constants to set $[\Psi]=0$. During the dissipative half-step we apply the resistive part of Ohm's law and the time-dependent form of Faraday's law \citep[Sec.~1]{laakmann2022augmented}. In each material they give
\[
\begin{aligned}
 E_{z,i}&=\eta\,\operatorname{curl}\vB_i
 =\eta(\partial_1 B_{i,2}-\partial_2 B_{i,1})=-\eta\nabla^2\Psi_i,\\
 \partial_t\vB_i&=-\operatorname{curl} E_{z,i}
 \quad\Longrightarrow\quad \operatorname{curl}(\partial_t\Psi_i+E_{z,i})=0 .
\end{aligned}
\]
The continuous traces of $\Psi$ and the tangential electric field $E_z$ give the same spatial constant $\partial_t\Psi_i+E_{z,i}$ on both sides. Subtracting its time integral from $\Psi$ gives
\[
 \partial_t\Psi_i=\eta\nabla^2\Psi_i\quad\text{in }H_i,\qquad
 [\Psi]=0,\quad [\gamma\partial_n\Psi]=0\quad\text{on }\Gamma .
\]

Let $K_{is}(x,y)$ be the response at $x\in H_i$ to a unit source at $y\in H_s$. Reflect $y$ across $\Gamma$ to $y^*$ and put $\ell=\sqrt{\eta\tau}$. On the source side add an image of strength $A_s$ to the direct Gaussian. Across the interface use a transmitted Gaussian of strength $C_s$,
\[
 Q=\Gamma_\ell(x-y),\qquad Q^*=\Gamma_\ell(x-y^*),\qquad
 K_{ss}=Q+A_sQ^*,\qquad K_{js}=C_sQ\quad(j\ne s).
\]
Each expression solves the heat equation in its half-plane, with the initial point mass only at $y$. On $\Gamma$, reflection gives $Q^*=Q$ and $\partial_nQ^*=-\partial_nQ$. Substituting into the two interface conditions gives
\[
 \begin{aligned}
 1+A_s&=C_s, & \gamma_s(1-A_s)&=\gamma_j C_s,\\
 A_s&=\kappa_s=\frac{\gamma_s-\gamma_j}{\gamma_s+\gamma_j}, & C_s&=1+\kappa_s .
 \end{aligned}
\]
For a target on side $i$, a source on the other side $j$ has $C_j=1+\kappa_j=1-\kappa_i$. The two target weights are
\begin{equation}
 K_{ii}(x,y)=Q+\kappa_i Q^*,\qquad K_{ij}(x,y)=(1-\kappa_i)Q,\qquad
 \kappa_f=\frac{\mu_b-1}{\mu_b+1},\quad \kappa_b=-\kappa_f .
\label{eq:ohmic-transmission}
\end{equation}
Reflection maps $H_i$ onto $H_j$, giving $J=\int_{H_i}Q^*\,dy=\int_{H_j}Q\,dy$. The kernel has unit mass, and subtracting the initial potential gives its half-step increment,
\[
\begin{aligned}
 \int_{H_i}K_{ii}\,dy+\int_{H_j}K_{ij}\,dy
   &=(1-J)+\kappa_iJ+(1-\kappa_i)J=1,\\
 \delta\Psi_i(x)=\Psi_i^+(x)-\Psi_i(x)
   &=\sum_{s\in\{f,b\}}\int_{H_s}K_{is}(x,y)\big[\Psi_s(y)-\Psi_i(x)\big]\,dy .
\end{aligned}
\]
For $\mu_b=1$ this reduces to free-space diffusion. At a permeability jump the same calculation fixes the reflected and transmitted contributions to the initial mesh weights.

\subsection{Ohmic Weights}
On the periodic mesh the uniform mean field $\vB_h$ has locally linear potential $\Psi_h$ and edge fluxes $b_h$. Integrating $b-b_h$ along a shortest-path spanning tree gives the periodic residual $\psi$ up to a root value,
\[
 \psi=\Psi-\Psi_h,\qquad G\psi=b-b_h,\qquad
 [\gamma\partial_n\psi]=-[\gamma\partial_n\Psi_h]=[\gamma]\,\vB_h\cdot\mathbf t .
\]
A tangential mean field at a permeability jump gives $\psi$ an inhomogeneous interface condition. We use the exact half-plane kernel for $\Psi$ in \eqref{eq:ohmic-transmission} as the initial shape of the weights for $\psi$.

With a separate bounded, tapered radial profile $m_\phi^B$, the image distance from \eqref{eq:image} and $\kappa_v$ from the target material, the raw cluster weights are
\[
\begin{gathered}
 k_P(r)=A_P\exp\!\left(-\frac{r^2}{4\ell^2}+m_\phi^B(r/\ell,\mathbf z)\right),
 \qquad Z_v=\sum_{Q\in\mathcal P_v}k_Q(r_{vQ}),\\
 \widetilde W_{vP}^{\,0}=\frac{1}{Z_v}
 \begin{cases}
 k_P(r_{vP})+\kappa_v k_P(r^*_{vP}),& v\text{ and }P\text{ on the same side},\\
 (1-\kappa_v)k_P(r_{vP}),& v\text{ and }P\text{ on opposite sides}.
 \end{cases}
\end{gathered}
\]
We apply the viscous bulk moment correction to $\widetilde W_{vP}^{\,0}$. On the mesh we use the signed distance $\varsigma$ to the fluid-body boundary, positive in the fluid. The tangential coordinate is stationary along a flat interface, as is $\zeta_i=\mu_i\varsigma$ because $[\zeta]=0$ and $\gamma_i\partial_{\varsigma}\zeta_i=1$. Table~\ref{tab:ohmic-moments} gives the inputs for each target node. Here $s^\parallel_{vP}$ is the centroid displacement along the nearest interface edge.

\begin{table}[!htbp]
\centering
\caption{Moment correction for the Ohmic exchange weights.}
\label{tab:ohmic-moments}
\small
\setlength{\tabcolsep}{9pt}
\begin{tabular*}{\linewidth}{@{\extracolsep{\fill}}lcc@{}}
\toprule
Location of node $v$ & First-moment coordinates & Second-moment input \\
\midrule
Bulk ($d_v>4\ell$) & $\mathbf s_P$ & $r_{vP}^2$ \\
Near the interface ($d_v\le4\ell$) & $(s^\parallel_{vP},\,\bar\zeta_P-\zeta_v)$ & $2(s^\parallel_{vP})^2$ \\
\bottomrule
\end{tabular*}
\end{table}
For a regular moment matrix, the zero first moment preserves both stationary potentials. Doubling the tangential variance $2\ell^2$ gives the bulk second-moment target $4\ell^2$.

With corrected weights $\widetilde W_{vP}$ and body indicator $\beta_v$, we sum the weighted differences of the cluster means $\bar\psi_P$ and nodal values $\psi_v$ in double precision. Oriented differences of the gated increment give the flux increments,
\[
\begin{aligned}
 g_v^{(\psi)}&=\operatorname{softplus}\!\left[G_\phi^B\!\left(
 \min(d_v/\ell,4),\beta_v,\operatorname{clip}(\log(h_v/\ell),-6,3),\mathbf z\right)\right],\\
 \delta\psi_v&=g_v^{(\psi)}\sum_{P\in\mathcal P_v}\widetilde W_{vP}(\bar\psi_P-\psi_v),\qquad
 (\delta b)_e=(G\,\delta\psi)_e=\delta\psi_{e^+}-\delta\psi_{e^-}.
\end{aligned}
\]
The Ohmic gate uses the same fixed bounds with the magnetic diffusion length $\ell=\sqrt{\eta\tau}$. Edges shorter than $e^{-6}\ell$ give the input $-6$, and edges longer than $e^3\ell$ give $3$. This bounds the gate input even when the local mesh spacing and the magnetic diffusion length differ by orders of magnitude. The direct and reflected kernels use the physical distances $r_{vP}$ and $r^*_{vP}$.
Each nodal increment occurs twice with opposite signs in the edge sum around a triangle. The sum vanishes, which preserves the discrete magnetic divergence.

Distillation trains the radial profile $m_\phi^B$ and the gate network $G_\phi^B$. We initialize their outputs to give
\[
 m_\phi^B\equiv0,\qquad g_v^{(\psi)}=1,\qquad k_P(r)=A_P\exp\!\left(-\frac{r^2}{4\ell^2}\right).
\]
The permeability fixes the reflection and transmission factors throughout training. Starting from this Gaussian exchange, the student fits the radial weights and nodal amplitudes to the teacher's edge-flux increments on \textit{permeable body}.

\section{Experimental Configurations}
\label{app:experiment-configs}

\subsection{Reference Data}
\label{app:reference-data}

The sweep contains 145 combinations of $(\nu,\eta,M_{s0},M_{A0})$. Each \textit{permeable body} operating point inherits the nominal $\mathrm{Re}_0$ and $\mathrm{Rm}_0$ levels from \textit{periodic box} ($2$, $5$, $12$, $28$ and $42$) and uses body permeabilities $\mu_b\in\{3,7.21,26.8\}$.

For \textit{permeable body} trajectories the total body area is $\pi(0.11)^2$ in the unit box. The mesh spacing is about $\nicefrac{1}{128}$ in the bulk and $\nicefrac{1}{256}$ along the body. The first layer at the wall has thickness between $1.5\cdot10^{-4}$ and $1.2\cdot10^{-3}$, depending on the estimated Hartmann layer thickness $\delta_H=\sqrt{\nu\eta}/|B_n|$ \citep{potherat2002effective}. To estimate $|B_n|$, we multiply the initial root mean square field by the largest wall-normal field induced by a unit applied field, taken over the body surface and the applied field direction. We select the largest available first-cell thickness $h_1$ for which the first three cells fit within the layer, using the finest available class if necessary.

For each single body, the radius at polar angle $\varphi$ is
\[
r(\varphi)=s\exp\!\left(\sum_{k=2}^{5}\big[a_k\cos(k\varphi)+b_k\sin(k\varphi)\big]\right),
\]
where $s$ fixes the area. The coefficients $a_k$ and $b_k$ are independent centered Gaussian draws with standard deviations $\sigma w_k$, where $w_k$ is proportional to $k^{-3/2}$ and $\sum_k w_k^2=1$. We draw $\sigma$ uniformly from $[0,0.25]$ and reject shapes outside the target $q$ band or with insufficient clearance for the boundary mesh layers.

Both solvers use the third-order implicit-explicit Runge--Kutta scheme ARS(4,4,3) \citep{ascher1997implicit}. The discretization for \textit{periodic box} uses $\nicefrac{3}{2}$ dealiasing and a maximum adaptive time step of $5\cdot10^{-3}$. The finite-element solver uses continuous P1 for density, continuous P2 for momentum, BDM$_2$ for the magnetic field and continuous P3 for the scalar electric field. Its adaptive time step is bounded by $2\cdot10^{-3}$. The stored \textit{permeable body} data include magnetic fluxes through the mesh edges.

Table~\ref{tab:dataset-quality} reports conservation and interface checks on the reference data. For each trajectory we record the maximum absolute mass drift relative to the initial mass and the maximum $\|\nabla\cdot\vB\|_{L^2}$ over the saved frames. We use Fourier derivatives on the periodic grid and the finite-element divergence on body-fitted meshes. From frame 5 onward we integrate viscous and Ohmic dissipation between consecutive frames by the trapezoidal rule. We pool the absolute differences from the observed free-energy drop, each divided by the magnitude of that drop.

At $t=0.2$ and $t=1$, we project the fields on either side of each interface vertex onto its normal and tangent. For each component we retain vertices where its fluid-side magnitude exceeds half its root mean square over the interface at that time. We compare the body-to-fluid ratio with $1$ for the normal component and with $\mu_b$ for the tangential component, then pool the absolute relative deviations across the retained vertices, frames and trajectories.

\begin{table}[!htbp]
\centering
\caption{Reference data checks. All reported values are 99th percentiles.}
\label{tab:dataset-quality}
\small
\setlength{\tabcolsep}{5pt}
\begin{tabular*}{\linewidth}{@{\extracolsep{\fill}}lcc@{}}
\toprule
Diagnostic & \textit{periodic box} & \textit{permeable body} \\
\midrule
Relative mass drift & $7.75\times10^{-10}$ & $1.43\times10^{-13}$ \\
Magnetic divergence ($L^2$ norm) & $8.60\times10^{-6}$ & $3.93\times10^{-12}$ \\
Free-energy balance (relative residual) & $0.50\%$ & $0.23\%$ \\
Normal magnetic field (relative jump error) & \tabledash & $2.84\%$ \\
Tangential magnetic field (relative jump error) & \tabledash & $13.60\%$ \\
\bottomrule
\end{tabular*}
\end{table}

\subsection{Model and Optimizer Settings}

Table~\ref{tab:periodic-configs} gives the configurations for the \textit{periodic box} ID comparison. The matched baseline widths give forward costs close to the $20.93$ GFLOP measured for \method{}. FNO acts directly on the $128\times128$ grid. Transolver uses periodic padding in its spatial projections. For MeshGraphNet the periodic triangulation connects nodes across opposite sides of the box. The baseline runs use seed zero and the same 15\,000-update pushforward protocol as \method{}.

\begin{table}[!htbp]
\centering
\caption{Model configurations for the \textit{periodic box} ID comparison. GFLOP counts refer to a forward step for one trajectory. The widths and costs of the three baselines describe the configurations used in Table~\ref{tab:periodic-id}.}
\label{tab:periodic-configs}
\small
\setlength{\tabcolsep}{4pt}
\begin{tabular*}{\linewidth}{@{\extracolsep{\fill}}l l r@{}}
\toprule
Model & Configuration & GFLOP \\
\midrule
\method{} & Four CNO levels, width multiplier 48 & 20.93 \\
FNO & Four Fourier layers, width 181, 32 modes per axis & 20.95 \\
Transolver & Eight layers, width 56, eight heads, 32 slices, MLP width 125 & 20.95 \\
MeshGraphNet & 15 message-passing steps, node and edge width 42 & 21.90 \\
\bottomrule
\end{tabular*}
\end{table}

The widths and optimizer settings for \textit{permeable body} are in Table~\ref{tab:baseline-configs}. Transolver uses eight layers, eight attention heads, 32 slices and an MLP expansion of two. MeshGraphNets uses 15 message-passing steps. GINO uses four Fourier layers with 32 modes per dimension on a $128\times128$ grid. The 2D volume adaptation of DoMINO uses one shared aggregation network. Its geometry radii are $(0.05,0.25,0.5,1.25)$ with neighbor limits $(32,64,128,256)$, and its local radii are $(0.05,0.125)$ with limits $(64,128)$.

\begin{table}[!htbp]
\centering
\caption{Model configurations for \textit{permeable body}. GINO widths list the Fourier width followed by the two hidden widths of the output graph kernel. The GFLOP column reports the cost of one forward step.}
\label{tab:baseline-configs}
\small
\setlength{\tabcolsep}{5pt}
\begin{tabular*}{\linewidth}{@{\extracolsep{\fill}}l l r l c c@{}}
\toprule
Model & Widths & GFLOP & Optimizer & Peak LR & Weight decay \\
\midrule
\method{} & 48 & 23.5 & Adam & $5\cdot10^{-4}$ & 0 \\
GINO & $(64,384,192)$ & 23.9 & AdamW & $10^{-3}$ & $10^{-4}$ \\
Transolver & 88 & 23.7 & Adam & $10^{-3}$ & 0 \\
MeshGraphNets & 37 & 23.7 & Adam & $10^{-4}$ & 0 \\
DoMINO & 168 & 24.0 & Adam & $10^{-3}$ & 0 \\
\midrule
GINO, $\times 2$ & $(96,576,288)$ & 46.5 & AdamW & $10^{-3}$ & $10^{-4}$ \\
GINO, $\times 4$ & $(144,864,432)$ & 96.0 & AdamW & $10^{-3}$ & $10^{-4}$ \\
GINO, $\times 8$ & $(208,1248,624)$ & 190.6 & AdamW & $10^{-3}$ & $10^{-4}$ \\
\bottomrule
\end{tabular*}
\end{table}

The three \textit{periodic box} models compared in Table~\ref{tab:periodic-rollout} use a four-level CNO with width multiplier 48, instance normalization and 16 channels at the input to the prediction head. The \textit{periodic box} implementation of \method{} conditions the trunk on $\log_{10}\nu$ and $\log_{10}\eta$ as well as the sonic and Alfv\'en Mach numbers. All three models use Adam with a peak learning rate of $5\cdot10^{-4}$ and zero weight decay. The increment-space metriplectic head uses width 128 and adds a degeneracy penalty and a magnetic-divergence penalty, both with weight $0.1$.

\bibliographystyle{elsarticle-num-names}
\bibliography{references}

\end{document}